# LiteInception: A Lightweight Interpretable Deep Learning Framework for General Aviation Fault Diagnosis

Zhihuan Wei [a], Xinhang Chen [a], Danyang Han [a], Yang Hu [a,*], Jie Liu [d], Xuewen Miao [b], Guijiang Li [c]

[a] *Hangzhou International Innovation Institute, Beihang University, Hangzhou, China*
[b] *Science and Technology on Complex Aviation System Simulation Laboratory, Beijing, China*
[c] *The First Aircraft Institute of Aviation Industry Corporation of China, Xi'an, China*
[d] *School of Reliability and Systems Engineering, Beihang University, Beijing, China*

**\* Corresponding author.** E-mail: yang_hu@buaa.edu.cn

## Highlights

• Two-stage cascaded architecture aligns with aviation maintenance workflows.

• 1+1 branch LiteInception achieves 70% compression with under 3% F1 loss.

• Knowledge distillation tunes precision–recall for different scenarios.

• Multi-method fusion selects 15 diagnostically valuable sensor channels.

• Dual-level interpretability localizes faulty sensors and time segments.

## Abatract

Fault diagnosis and efficient maintenance of general aviation (GA) aircraft are critical for flight safety and economic operations. However, deploying deep learning models on resource-constrained edge devices poses dual challenges of computational capacity and interpretability. This paper proposes LiteInception—a lightweight and interpretable fault diagnosis framework designed for edge deployment. The framework adopts a two-stage cascaded architecture aligned with standard maintenance workflows: the first stage performs high-recall fault detection, and the second stage conducts fine-grained fault classification on anomalous samples, thereby achieving decoupled optimization objectives and on-demand allocation of computational resources. For model compression, a multi-method fusion strategy based on mutual information, gradient analysis, and Squeeze-and-Excitation (SE) attention weights is proposed to reduce the input sensor channels from 23 to 15 dimensions. A 1+1 branch LiteInception architecture is further proposed, compressing the parameter count of InceptionTime by 70%, accelerating CPU inference by over 8×, with a performance loss of only 3%. Moreover, knowledge distillation is introduced as a precision–recall trade-off mechanism, enabling the same lightweight model to bias toward either high recall or high precision simply by switching the distillation training strategy, thereby adapting to different deployment scenarios such as safety-critical applications (prioritizing missed-detection avoidance) and assisted diagnosis (prioritizing false-alarm reduction). Finally, a dual-level interpretability framework integrating four attribution methods—Input Gradient, Occlusion Sensitivity, Grad-CAM, and Integrated Gradients—is constructed to localize suspected faulty sensors among the 15 input channels and identify critical anomalous temporal segments within 2,048-time-step flight records, providing maintenance personnel with a traceable evidence chain of "which sensor × which time segment." Experiments on the NGAFID aviation maintenance dataset demonstrate that the proposed method achieves a fault detection accuracy of 81.92%, a recall of 83.24%, and a fault identification accuracy of 77.00%, validating a favorable balance among efficiency, accuracy, and interpretability.

# 1 Introduction

Fault diagnosis of general aviation aircraft is critical to flight safety. Studies have shown that the accident rate in general aviation is significantly higher than that in commercial aviation, with aircraft characteristics (including mechanical failures and engine malfunctions) constituting the second most important causal category after human factors [1], [2], a considerable proportion of which could have been avoided through timely and effective fault diagnosis. As a core component of Prognostics and Health Management (PHM), fault diagnosis plays a vital role in reducing life-cycle maintenance costs and ensuring flight safety through proactive monitoring and condition assessment of critical subsystems [3]. However, real-world general aviation flight recording data, as exemplified by the NGAFID dataset, pose formidable challenges to existing methods: GA aircraft are equipped only with basic flight and engine sensors rather than dedicated fault monitoring instruments, and the variability introduced by external factors such as pilot maneuvers, weather conditions, and mission profiles far exceeds the subtle feature shifts caused by component degradation. Moreover, chronic wear-type faults predominate, resulting in extremely subtle differences in sensor readings before and after fault onset [4]. These data characteristics—high noise, strong interference, and weak fault signatures—render conventional statistical threshold-based detection methods and unsupervised approaches such as autoencoders insufficient for effectively distinguishing genuine faults from normal flight variations. In recent years, deep learning methods have achieved remarkable progress in time series analysis and fault diagnosis [5], offering new technical avenues for addressing the aforementioned data challenges. Nevertheless, applying these methods to real-world GA aircraft maintenance scenarios still faces several core challenges.

First, the resource constraints of edge deployment are prominent. GA aircraft require diagnostic decisions to be completed during ground turnaround periods (typically 30 to 60 minutes); however, the computational capacity of onboard embedded devices or ground-based edge terminals is far inferior to that of server-grade GPUs. Existing deep learning architectures have excessively large parameter counts, making it difficult to meet real-time inference requirements, and exploration of lightweight methods tailored to aviation time series fault diagnosis tasks remains relatively limited. Second, there is a disconnect between actual maintenance workflows and model design. Real-world aviation maintenance follows a hierarchical process of "monitoring → alerting → diagnosis → decision-making," whereas most existing methods adopt single-stage end-to-end designs that mix normal samples with fault types for multi-class classification. This not only leads to computational waste (normal samples typically exceed 80%) but also makes it difficult for a single model to simultaneously optimize fault missed detection (high safety cost) and false alarms (low monetary cost)—two objectives with severely asymmetric costs. Third, high-dimensional sensor data contain redundancy and noise interference. GA aircraft are equipped with dozens of sensor channels, but not all channels contribute equally to fault diagnosis. Traditional feature engineering methods rely on domain expert experience for channel selection [6], which is time-consuming and difficult to adapt to varying operating conditions. Fourth, the demand for interpretability constrains the application of deep learning in aviation safety domains. Maintenance personnel need to know "which sensor" exhibited anomalies during "which time segment" to efficiently localize faults; black-box predictions cannot satisfy the compliance requirements of airworthiness certification, and existing interpretability methods are mostly designed for image data, with their applicability to multivariate time series data yet to be validated. Finally, there is a lack of flexible mechanisms for tuning the trade-off between precision and recall. Safety-critical scenarios require high recall to avoid missed detections, whereas assisted diagnosis scenarios demand high precision to reduce false alarms. Existing methods typically yield a single model through a fixed training strategy and lack the means to flexibly balance these two metrics without modifying the network architecture. Although knowledge distillation has been widely used for model compression [7], its potential for modulating the precision–recall trade-off has not yet been explored.

To address the aforementioned challenges, this paper proposes LiteInception—a lightweight and interpretable deep learning framework for fault diagnosis of general aviation aircraft. The main contributions of this paper are as follows:

(1) Two-stage cascaded diagnosis architecture. A "fault detection + fault identification" cascaded architecture aligned with standard maintenance workflows is designed: the first stage performs high-recall fault detection, and the second stage conducts fine-grained fault type identification on anomalous samples, achieving decoupled optimization objectives and on-demand allocation of computational resources.

(2) Multi-method fusion sensor channel selection strategy. A multi-method fusion channel selection approach based on mutual information, gradient analysis, and SE attention weights is proposed, reducing the number of sensor channels from 23 to 15 while improving diagnostic performance and effectively eliminating redundancy and noise interference.

(3) LiteInception lightweight architecture. A 1+1 branch architecture is designed, compressing the parameter count of InceptionTime by 70%, accelerating CPU inference by over 8×, and controlling performance loss to within 3%.

(4) Precision–recall tuning via knowledge distillation. A knowledge distillation training strategy is introduced, enabling the same lightweight model to flexibly adjust between precision and recall simply by switching the training approach, thereby adapting to different scenario requirements such as safety-critical and assisted diagnosis applications.

(5) Dual-level interpretability analysis framework. A dual-level interpretability framework at the sensor level and temporal segment level is constructed, integrating four complementary attribution methods to provide a traceable evidence chain of "which sensor × which time segment" for fault diagnosis results.

Experiments on the real-world NGAFID aviation maintenance dataset [8] demonstrate that the proposed method achieves a favorable balance among efficiency, accuracy, and interpretability.

The remainder of this paper is organized as follows: Section 2 reviews related work; Section 3 details the proposed method; Section 4 reports experimental results; Section 5 provides discussion and analysis; and Section 6 concludes the paper.

## 2 Related Work

This study spans five intersecting fields: aviation fault diagnosis, deep learning for time series classification, sensor feature selection, model compression and knowledge distillation, and explainable artificial intelligence. Significant progress has been made in each of these areas in recent years; however, existing work still exhibits clear shortcomings when targeting fault diagnosis scenarios for edge deployment on general aviation aircraft. The following provides a detailed review organized along these five dimensions.

Aircraft fault diagnosis is a core research direction within Prognostics and Health Management (PHM) [3] . Data-driven methods have progressively become the dominant paradigm: early approaches based on Support Vector Machines (SVM) [9] and Random Forests demonstrated feasibility on specific tasks, followed by deep learning methods—represented by Convolutional Neural Networks (CNNs) and Recurrent Neural Networks (RNNs)—that exhibited stronger capabilities for automatic fault feature extraction and classification [10]. Xue et al. [6] provided a systematic review of machine learning modeling approaches for gas turbines, identifying insufficient data availability and feature engineering quality as key bottlenecks constraining diagnostic performance while surveying feature selection and feature learning techniques. Khan et al. [11] reviewed deep learning applications in system health management, covering multiple engineering systems including aero-engines, and revealed the potential of deep models such as autoencoders and convolutional networks for unsupervised anomaly detection and fault prediction. In the aviation domain, data-driven methods have been widely applied to engine health management and remaining useful life (RUL) prediction, encompassing diverse technical approaches including multi-cell LSTM [12] , dual-attention mechanisms [13] , data–physics fusion [14] , and semi-supervised deep architectures [15] .

Nevertheless, existing methods exhibit two notable deficiencies. First, the vast majority adopt a single-stage end-to-end classification paradigm that mixes normal samples (typically exceeding 95% in actual operations) with multiple fault types for multi-class classification, performing full multi-class inference on all samples and thereby incurring substantial unnecessary computational overhead. Second, the severe asymmetry in costs—where missed fault detection may lead to safety incidents while false alarms merely increase re-inspection costs—makes it difficult for a single model to simultaneously customize for these two distinct optimization objectives. Cascaded or hierarchical diagnostic architectures have been preliminarily explored in industrial settings: Zhang et al. [16] proposed a two-stage method integrating optimized SVDD with SVM, and Gan et al. [17] constructed a hierarchical diagnostic network based on Deep Belief Networks. However, these works are mostly designed for vibration signals of individual equipment, neither considering stage-wise optimization of precision and recall nor aligning with actual maintenance workflows.

In the field of time series classification, deep learning methods have demonstrated powerful modeling capabilities. Wang et al. [18] first proposed end-to-end baselines based on Fully Convolutional Networks (FCN) and Residual Networks (ResNet), validating the effectiveness of one-dimensional convolutional architectures on the UCR benchmark datasets. Ismail Fawaz et al. [19] proposed InceptionTime, which simultaneously captures multi-scale temporal patterns through a parallel structure comprising three groups of convolutional branches with different kernel sizes and a MaxPool branch, achieving classification accuracy on par with the then state-of-the-art methods while offering far superior scalability. Subsequently, MiniRocket [20] achieved extremely high training efficiency through fixed convolutional kernel transformations. Transformer architectures have also been introduced into time series modeling: Informer [21] reduced the computational complexity for long sequences via sparse attention, and TimesNet [22] leveraged periodicity to reshape one-dimensional time series into two-dimensional tensors for feature extraction. Additionally, multi-scale Inception structures combining convolution with Gated Linear Units [23], the Temporal Fusion Transformer [24], and tightly coupled convolution–Transformer models [25] have all demonstrated competitiveness across various time series tasks. Zhao et al. [26] provided a systematic review of deep learning applications in machine health monitoring, covering the performance of architectures such as autoencoders, CNNs, and RNNs in time series signal processing. However, the aforementioned methods face clear bottlenecks for edge deployment. InceptionTime employs an ensemble strategy of five networks with internal multi-branch parallel structures, resulting in a high total parameter count and computational cost. Transformer-based models have even larger parameter counts and often underperform convolutional architectures in scenarios with limited training data or signals exhibiting distinct local patterns [27]. Overall, InceptionTime demonstrates a favorable balance between time series classification accuracy and local feature adaptability, but systematic lightweight transformation of its multi-branch parallel structure remains unexplored.

In multi-channel time series fault diagnosis, sensor channel selection is a critical step for improving model performance and reducing computational overhead. Existing methods can be broadly categorized into three classes. Filter-based methods employ statistical measures to evaluate and rank features prior to model training. For example, Verron et al. [28] proposed a mutual information-based feature selection method. Such methods are computationally efficient but evaluate features independently of the downstream classifier, failing to reflect how features are actually utilized within deep learning models. Wrapper-based methods directly use classifier performance as the evaluation criterion. For instance, He et al. [29] adopted a two-stage strategy combining mutual information pre-screening with multi-objective genetic algorithms. These methods typically achieve higher classification accuracy but suffer from computational costs that escalate sharply with feature dimensionality. Embedded methods integrate channel selection into the model training process, among which channel attention mechanisms represented by the SE (Squeeze-and-Excitation) module [30] have been widely applied to fault diagnosis tasks [31], adaptively assigning importance weights to each channel. However, attention weights may lack stability due to training stochasticity. Given that each of the three categories has its own shortcomings, using any single approach alone is unlikely to yield robust and reliable channel selection results. Moreover, existing studies predominantly focus on vibration signals from industrial rotating machinery [31], whereas aviation flight sensor systems comprise heterogeneous signals spanning diverse physical dimensions such as temperature, pressure, and electrical parameters [32]. The channel selection problem in this context faces unique challenges including signal type diversity and significant physical scale disparities, and relevant research remains scarce.

Model compression is a key enabling technology for edge deployment. In the image domain, MobileNet's depthwise separable convolutions [33], ShuffleNet's channel shuffling [34], and EfficientNet's neural architecture search [35] have each achieved notable compression results. However, these methods are all premised on two-dimensional spatial locality assumptions, which differ structurally from one-dimensional time series data, limiting the effectiveness of direct transfer. Lightweight research in the time series domain has been relatively lagging: Yan et al. [36] proposed the LiConvFormer framework, designing separable multi-scale convolutional modules that substantially reduce parameters and floating-point operations while maintaining diagnostic performance. Li et al. [37] constructed the DPW-ATTCNN lightweight model based on depthwise separable convolutions and efficient channel attention. However, these works focus on Transformer hybrid architectures or single-branch convolutional networks, respectively. More recently, Ismail-Fawaz et al. [38] proposed the LITE architecture, which achieves parameter compression by replacing the standard convolutions within Inception modules with depthwise separable convolutions. Nevertheless, LITE retains the original multi-branch topology, and its channel-independence assumption may sever the physical coupling among heterogeneous sensors. Systematic pruning of the multi-branch structure at the inter-branch topological level has yet to be explored.

Regarding knowledge distillation, Hinton et al. [7] proposed the classical teacher–student framework, achieving knowledge transfer by having the student model learn soft labels from the teacher's outputs. Romero et al. [39]proposed FitNets, further guiding the student to learn intermediate-layer representations from the teacher. Existing studies have demonstrated that soft-label learning enables models to obtain smoother decision boundaries [40], and Tao et al. [41] further revealed the differentiated impact mechanisms of samples with varying difficulty levels during the distillation process. However, existing distillation research has focused almost exclusively on maintaining accuracy while compressing models, neglecting the possibility that the distillation process may systematically alter the model's precision–recall trade-off characteristics. This property holds unique value in aviation fault diagnosis—where different scenarios exhibit markedly different tolerances for missed detections versus false alarms—yet it has not been systematically explored or exploited.

Explainable Artificial Intelligence (XAI) aims to reveal the decision-making mechanisms of deep learning models to meet compliance requirements in safety-critical domains. Existing methods can be broadly classified into three categories: gradient-based methods, such as the Input Gradient method of Simonyan et al. [42] and the Integrated Gradients method of Sundararajan et al. [43]; class activation mapping-based methods, such as CAM [44] and Grad-CAM [45]; and perturbation-based methods, such as the Occlusion Sensitivity analysis of Zeiler et al. [46] and the SHAP method of Lundberg et al. [47] based on Shapley values. In the fault diagnosis domain, interpretability methods have seen preliminary application. Zhao et al. [48] utilized signal-to-image mapping combined with Grad-CAM to visualize the regions of bearing fault features attended to by CNNs. Brito et al. [49] proposed a post-hoc explanation scheme based on unsupervised anomaly detection and Grad-CAM. Tang et al. [50] constructed an interpretable fault diagnosis framework based on LightGBM and TreeSHAP. Meng et al. [51] explored the application of Shapley value-based feature contribution quantification methods in continuous monitoring of rotating machinery. However, existing work exhibits two clear limitations. First, most studies employ only a single interpretability method, lacking multi-method cross-validation to assess the robustness of interpretation results. Second, existing methods are mostly designed for image or univariate signals, with insufficient attention to the joint interpretation across the sensor channel dimension and the temporal dimension of multivariate time series data. In aviation fault diagnosis, maintenance personnel need to know not only "when" an anomaly occurred but also "which sensor" emitted the anomalous signal.

## 3 Methodology

### 3.1 Problem Definition and Framework Overview

Fault diagnosis of general aviation aircraft can be formalized as a multivariate time series classification problem. Let the flight sensor dataset be $D = (X_i, y_i)_{i=1}^{N}$, where $N$ is the total number of samples. Each input sample $X \in \mathbb{R}^{T \times C}$ represents a multi-sensor channel recording of a flight phase, where $T$ is the time step length ($T = 2048$) and $C$ is the number of sensor channels after channel selection ($C = 15$). The classification label $y \in 1, 2, \ldots, K$ denotes the fault category. The raw data undergo missing value linear interpolation, resampling alignment, and Min-Max normalization preprocessing to ensure temporal consistency and numerical stability of the inputs.

In accordance with the hierarchical workflow of aviation maintenance—"monitoring → alerting → diagnosis → decision-making"—this paper decomposes the above classification problem into two cascaded subtasks. The first stage is fault detection as a binary classification task, with the objective of learning the mapping $f^{(1)} : \mathbb{R}^{T \times C} \to 0, 1$ ($0$ : normal, $1$ : anomalous). The second stage is fault identification as a multi-class classification task, with the objective of learning the mapping $f^{(2)} : \mathbb{R}^{T \times C} \to 1, 2, \ldots, K$ ($K = 19$, covering typical fault modes such as engine running abnormalities, seal damage, and baffle faults). The overall decision function of the two-stage cascaded diagnostic system can be expressed as:

$$\hat{y} = \begin{cases} 0, f^{(1)}(X) = 0 \\ f^{(2)}(X), f^{(1)}(X) = 1 \end{cases} \quad (1)$$

That is, the second-stage fine-grained fault identification is triggered only when the first stage determines the sample to be anomalous.

The overall methodological workflow of the LiteInception framework proceeds as follows: first, a multi-method fusion sensor channel selection strategy is applied to select the 15 most diagnostically valuable channels from the original 23 sensor channels (Section 3.2); second, a 1+1 branch LiteInception lightweight architecture is designed as the core feature extraction module (Section 3.3); then, a two-stage cascaded diagnostic system is constructed based on this module (Section 3.4); subsequently, knowledge distillation is introduced as a precision–recall tuning mechanism to adapt to different deployment scenarios (Section 3.5); finally, a dual-level interpretability framework is constructed to provide evidentiary support for diagnostic conclusions (Section 3.6).

## 3.2 Multi-Method Fusion Sensor Channel Selection

Multi-channel time series data often contain redundant or noisy sensor channels that not only increase computational burden but may also introduce spurious correlations leading to overfitting. The core task of sensor channel selection is to identify, from the original channel set, the subset that contributes most to fault diagnosis: retaining high-value channels that are strongly correlated with fault categories and actually relied upon by the model, while excluding low-value channels that are unrelated to faults or carry redundant information. However, each single feature evaluation method has its own limitations—statistical methods cannot reflect the model's actual learning behavior, while model-based methods may be affected by training stochasticity. To this end, this paper proposes a sensor channel selection strategy that fuses three complementary perspectives, enhancing the robustness of selection results through cross-validation.

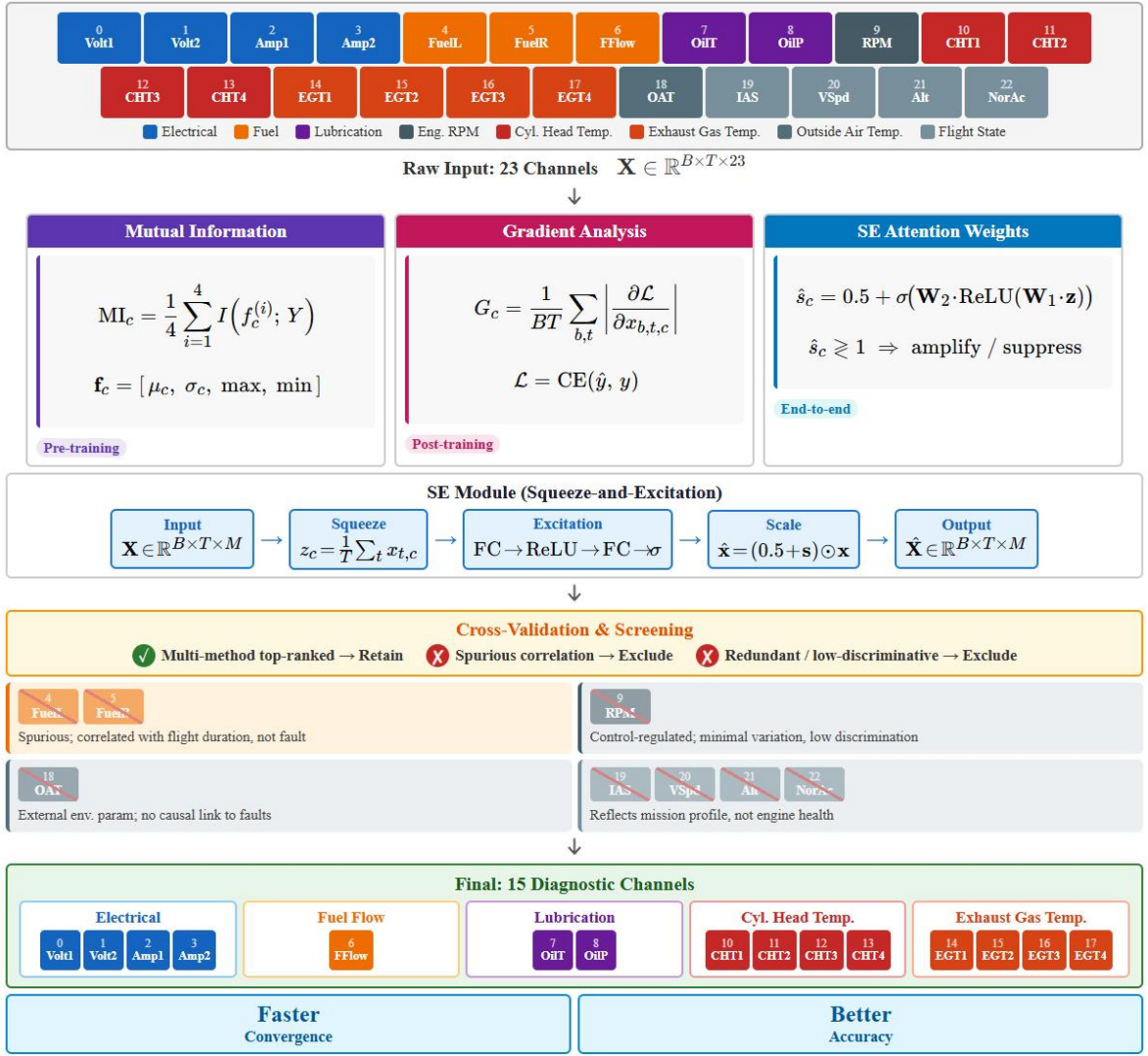

**Figure 1.** Sensor channel selection architecture.

### 3.2.1 Sensor Selection Methods

Mutual Information (MI) evaluates the statistical dependency between each channel and the fault labels prior to model training. Due to the excessively high dimensionality of the raw time series, this paper extracts a statistical feature vector $f_c = [\mu_c, \sigma_c, \max(x_c), \min(x_c)]$ from the time series $x_c \in \mathbb{R}^T$ of each sensor channel $c$. The MI score for sensor channel $c$ is:

$$MI_c = \frac{1}{4}\sum_{i=1}^{4} I(f_c^{(i)}; Y) \tag{2}$$

where $I(\cdot;\cdot)$ is the mutual information function and $Y$ is the fault category label. A higher MI score indicates a stronger dependency between the channel's statistical features and the fault categories, meaning the information carried by that channel is more valuable for distinguishing different fault states and should be preferentially retained. Channels with MI scores close to zero are approximately statistically independent of the fault categories and are candidates for exclusion. This method is computationally efficient and model-agnostic; however, its limitation lies in measuring only marginal statistical correlations, unable to reflect how features are actually utilized within the nonlinear mappings of deep learning models.

Gradient analysis quantifies the contribution of each channel's input to the classification loss via backpropagation, directly reflecting feature dependencies during the model's learning process. For input $X \in \mathbb{R}^{B \times T \times M}$ ($B$ is the batch size, $M$ is the number of channels) and cross-entropy loss $L$, the gradient importance of sensor channel $c$ is defined as:

$$G_c = \frac{1}{B \cdot T} \sum_{b=1}^{B} \sum_{t=1}^{T} \left| \frac{\partial L}{\partial x_{b,t,c}} \right| \tag{3}$$

Aggregation of gradient absolute values eliminates the cancellation effect of positive and negative gradients. A higher gradient importance indicates that small perturbations to the channel's input have a greater impact on the classification loss, meaning the model relies more heavily on that channel when making diagnostic decisions and it should be retained. Channels with gradient importance close to zero imply that the model has extracted virtually no effective discriminative information from them and are candidates for exclusion. This method directly reflects the model's actual learning behavior but may be influenced by specific training stages and parameter initialization, with rankings potentially fluctuating across different training epochs.

The Squeeze-and-Excitation (SE) module assigns adaptive weights to each channel through end-to-end learning, providing an optimization-driven perspective for channel evaluation. The SE module first performs global average pooling along the temporal dimension to obtain a channel descriptor $z \in \mathbb{R}^M$:

$$z_c = \frac{1}{T}\sum_{t=1}^{T} x_{t,c} \tag{4}$$

Then, channel weights are learned through a two-layer fully connected network:

$$s = \sigma(W_2 \cdot \text{ReLU}(W_1 \cdot z)) \tag{5}$$

where $W_1 \in \mathbb{R}^{(M/r) \times M}$, $W_2 \in \mathbb{R}^{M \times (M/r)}$, $r$ is the reduction ratio, and $\sigma$ is the Sigmoid function. To prevent information loss due to excessively small weights, $s$ is linearly transformed to the interval $[0.5, 1.5]$: $\hat{s}_c = 0.5 + s_c$. A weight greater than 1 indicates that the model actively amplifies the feature response of that channel during optimization, suggesting a positive contribution to the classification objective and warranting retention. A weight less than 1 indicates that the model suppresses that channel, suggesting limited informational value or the presence of interference, making it a candidate for exclusion. Unlike the previous two methods, SE weights are implicitly learned during end-to-end optimization toward the classification objective, reflecting the model's global assessment of channel importance.

### 3.2.2 Multi-Method Cross-Validation and Fusion Decision

The three methods described above provide importance scores for each channel from three distinct dimensions: statistical correlation (MI: dependency strength between channel and label), model sensitivity (gradient: degree of channel influence on classification loss), and optimization preference (SE weight: degree to which the model amplifies or suppresses the channel). Each method, considered independently, may produce biases—MI may overestimate channels with spurious correlations to labels, gradient rankings may fluctuate due to training stochasticity, and SE weights depend on the specific network architecture. This paper adopts a multi-method cross-validation fusion strategy: first, sensor channel importance rankings are obtained independently from each method; then, for each sensor channel, the consistency of its performance across the three methods is comprehensively analyzed, and final decisions are made in conjunction with domain physical knowledge. The specific decision criteria are as follows:

(1) Multi-method consistency criterion: If a sensor channel ranks highly across all three methods, it is deemed to possess robust diagnostic value and is retained. If a sensor channel ranks highly in only one method while performing mediocrely in the others, its high ranking may stem from statistical coincidence or spurious correlations, and it should be excluded.

(2) Physical causality criterion: Even if a sensor channel performs well on statistical metrics, it should be excluded if it lacks a plausible physical causal relationship with the fault (e.g., parameters reflecting flight mission characteristics rather than engine health status) or if the causal direction is erroneous (e.g., a consequence rather than a cause of the fault).

(3) Information redundancy criterion: If two sensor channels carry highly correlated information, the channel more directly related to the fault mechanism and richer in informational content is preferentially retained, while the redundant channel is excluded.

The specific sensor channel selection results and their physical interpretations are detailed in Section 4.2.

### 3.3 LiteInception Lightweight Architecture

This section proposes LiteInception—a lightweight time series feature extraction architecture designed for edge deployment. Building upon InceptionTime, this architecture achieves substantial parameter and computational compression through systematic branch pruning while maintaining performance loss within an acceptable range.

#### 3.3.1 InceptionTime Architecture Review

The core of InceptionTime is the Inception module, which employs a 3+1 branch parallel structure: three convolutional branches with different kernel sizes ($k_1, k_2, k_3$) simultaneously capture multi-scale temporal patterns, augmented by a MaxPool branch that extracts local extremum features. Let the input feature map dimension be $\mathbb{R}^{T' \times D_{in}}$; each convolutional branch outputs $D_f$ feature channels, and the MaxPool branch likewise outputs $D_f$ feature channels, yielding a single Inception module output dimension of $\mathbb{R}^{T' \times 4D_f}$.

Under the standard configuration with a bottleneck layer, the parameter count of a single Inception module can be decomposed as:

$$P_{3+1} = D_{in} \cdot D_b + D_b \cdot D_f \cdot (k_1 + k_2 + k_3) + D_{in} \cdot D_f \tag{6}$$

where $D_b$ is the bottleneck dimension, the first term represents bottleneck layer parameters, the middle term represents parameters of the three convolutional branches, and the last term represents parameters of the $1\times1$ convolution following MaxPool. Taking the standard configuration $k_1 = 3, k_2 = 5, k_3 = 7$, $D_f = 128$, $D_b = 64$ as an example, the three convolutional branches account for approximately 75% of the total module parameters.

### 3.3.2 Motivation for Lightweight Design

The design decision to prune the 3+1 branch structure to a 1+1 branch structure is based on the following theoretical analysis.

Receptive field redundancy analysis. In a 6-layer stacked structure, a single $k = 3$ convolutional branch achieves a theoretical receptive field of $1 + 6 \times (3-1) = 13$ time steps after layer-by-layer stacking, whereas in the original 3+1 structure, the $k = 7$ branch has a single-layer receptive field of 7, reaching $1 + 6 \times (7-1) = 37$ after 6 layers. For aviation sensor data resampled to 2,048 steps, local fault signatures typically manifest as short-duration amplitude anomalies or trend deviations in specific frequency bands, with characteristic scales concentrated in the range of tens to over a hundred time steps. A $k = 3$ convolution can cover this range through 6-layer stacking, and layer-by-layer stacking provides stronger nonlinear expressiveness compared to single-layer large kernels—each convolutional layer is followed by batch normalization and ReLU activation, introducing multiple nonlinear transformations, whereas large-kernel convolutions perform only a single linear mapping followed by activation. Therefore, the additional receptive field provided by the $k=5$ and $k=7$ branches in the multi-scale parallel structure exhibits significant redundancy.

Indispensability of the MaxPool branch. The MaxPool operation extracts the maximum value within a sliding window, making it naturally suited for capturing extremum features in sensor signals such as peak anomalies and threshold exceedances. These features carry direct physical significance in fault diagnosis (e.g., temperature spikes, pressure drops). Convolution operations extract trend and pattern features through weighted summation, complementing the extremum extraction of MaxPool. This complementarity implies that the information gain from the MaxPool branch cannot be fully substituted by convolutional branches.

Distinction from lightweight methods in the visual domain. MobileNet's depthwise separable convolutions and ShuffleNet's channel shuffling are both designed for two-dimensional spatial locality, with the core assumption that the two spatial dimensions possess symmetry. However, one-dimensional time series data have only a temporal dimension; the channel-independence assumption of depthwise separable convolutions severs the physical coupling between sensors (e.g., the linkage between cylinder head temperature and exhaust gas temperature), making direct transfer inappropriate. The strategy adopted in this paper is to achieve lightweight design by reducing the number of parallel branches while preserving the full cross-channel interaction capability of standard convolutions, which is better suited to the characteristics of multi-channel time series fault diagnosis tasks.

### 3.3.3 1+1 Branch LiteInception Architecture

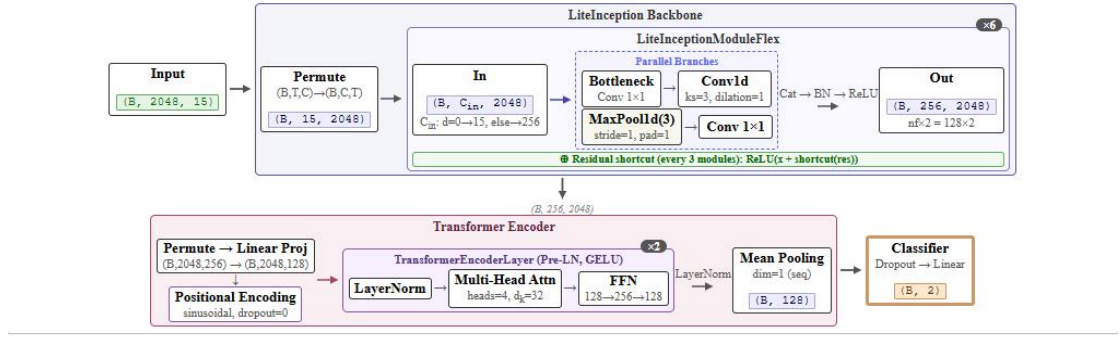
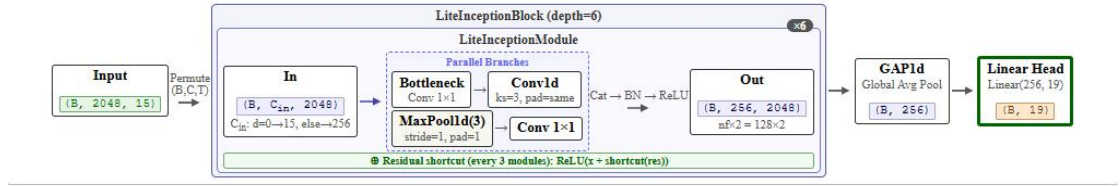

**Figure 2.** Model architecture diagram.

Based on the above analysis, LiteInception prunes the original 3+1 parallel branches to a 1+1 structure: retaining the combination of a single $k=3$ convolutional branch and a MaxPool branch. The structure of a single LiteInception module is as follows:

Convolutional branch: The input first passes through a $1\times1$ bottleneck convolution to compress the feature channel count from $D_{in}$ to $D_b$, then through a $k=3$ one-dimensional convolution to map to $D_f$ output channels:

$$H_{conv} = \text{Conv1d}_{k=3}(\text{Conv1d}_{1\times1}(X)) \qquad (7)$$

MaxPool branch: The input passes through max pooling with kernel size 3 and stride 1, followed by a $1\times1$ convolution to adjust the feature channel count to $D_f$:

$$H_{pool} = \text{Conv1d}_{1\times1}(\text{MaxPool}_{k=3}(X)) \qquad (8)$$

Feature fusion: The outputs of both branches are concatenated along the feature channel dimension, followed by batch normalization and ReLU activation:

$$H_{out} = \text{ReLU}(\text{BN}(\text{Concat}(H_{conv}, H_{pool}))) \qquad (9)$$

The output dimension is $\mathbb{R}^{T' \times 2D_f}$. Batch normalization is placed before the activation function (Conv→BN→ReLU) to normalize the feature distribution prior to the nonlinear transformation.

The parameter count of a LiteInception module is:

$$P_{1+1} = D_{in} \cdot D_b + D_b \cdot D_f \cdot k + D_{in} \cdot D_f \tag{10}$$

Compared to the original 3+1 structure, the convolutional branch parameters are reduced from $D_b \cdot D_f \cdot (k_1 + k_2 + k_3)$ to $D_b \cdot D_f \cdot k$. Taking $D_f = 128$, $D_b = 64$, $k = 3$ as an example, the per-module parameter count is reduced from approximately 135K to approximately 41K, a compression ratio of approximately 3.3×.

The complete LiteInception network consists of 6 stacked LiteInception modules, with residual connections placed between every 3 modules to mitigate gradient degradation. At the network's end, Global Average Pooling (GAP) aggregates temporal features into a fixed-length vector representation, followed by a fully connected layer that outputs the classification result. The total network parameter count is approximately 639K, representing a 70.4% compression from InceptionTime's 2,159K; FLOPs are reduced from 8.77G to 2.58G, a 70.6% reduction in computational cost.

### 3.4 Two-Stage Cascaded Diagnostic Architecture

Conventional single-stage end-to-end methods mix normal samples with fault types for multi-class classification, suffering from two limitations. First, in real operational environments, the majority of aircraft flights are in a normal state, and performing full multi-class inference on all samples results in computational waste. Second, the severely asymmetric costs of missed fault detection (high safety cost) versus false alarms (low monetary cost) make it difficult for a single model to simultaneously optimize both objectives. This paper proposes a two-stage cascaded architecture aligned with the maintenance workflow, decoupling "whether anomalous" and "what type of fault" into two independently optimized subtasks.

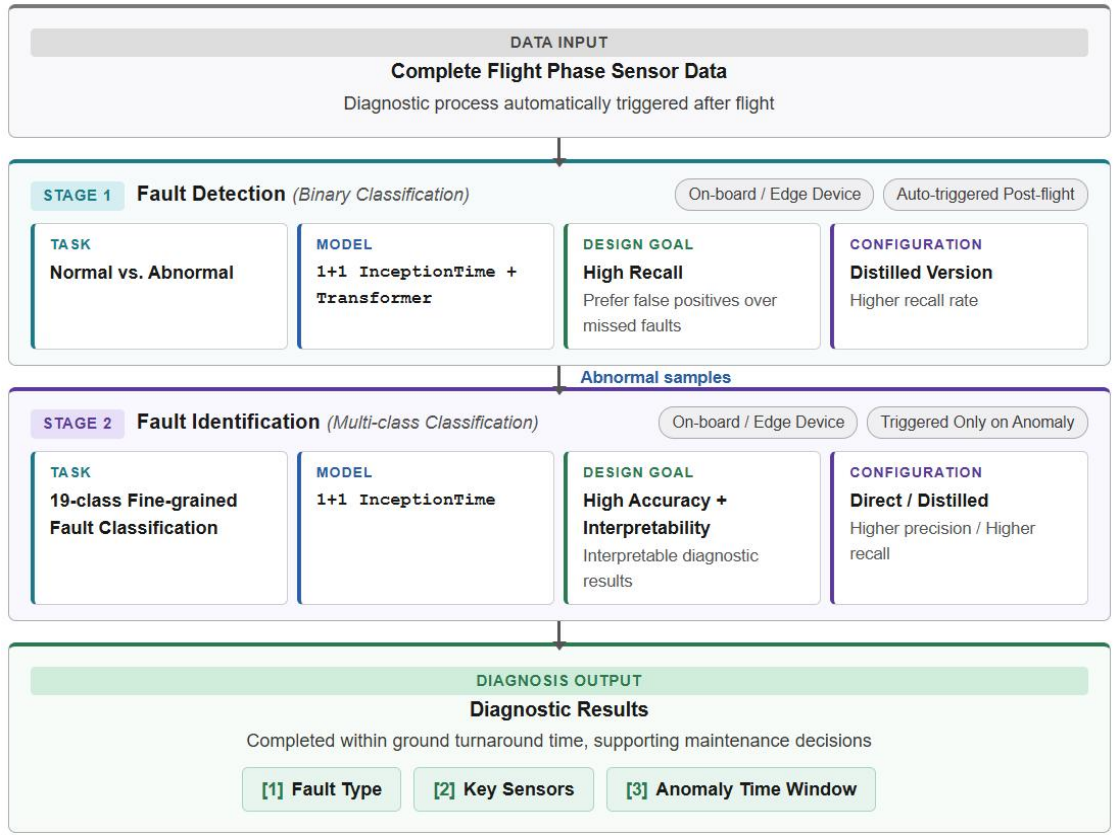

**Figure 3.** Two-stage diagnostic architecture schematic.

### 3.4.1 Stage Model Configuration

Stage 1 (Fault Detection) employs a hybrid architecture combining LiteInception with a Transformer encoder. The motivation for the hybrid architecture is as follows: fault detection, as a binary classification task, is less discriminatively challenging than 19-class fine-grained classification, but is more sensitive to the capture of global temporal patterns—certain anomalies may manifest only as overall parameter shifts during specific segments in the mid-to-late phases of a flight, rather than local waveform abnormalities. LiteInception is responsible for extracting local temporal features, while the Transformer encoder captures long-range temporal dependencies through the self-attention mechanism, and the two are complementary. The optimization objective of this stage is to maximize recall, adhering to the safety principle of "better a false alarm than a missed detection."

Stage 2 (Fault Identification) employs a pure LiteInception architecture without introducing a Transformer module. The rationale is as follows: fine-grained discrimination among 19 fault classes relies primarily on precise delineation of inter-class boundaries in the feature space rather than long-range dependency modeling. Furthermore, the second stage processes only the subset of anomalous samples, resulting in a smaller data volume, where the large parameter count of Transformers is instead prone to overfitting (this point will be corroborated by the experimental results in Section 4.3, where the hybrid architecture containing a Transformer substantially underperforms the pure convolutional architecture on the 19-class task). The optimization objective of this stage is high accuracy and interpretability.

### 3.4.2 Architectural Advantages

The two-stage design yields three core advantages. First, objective decoupling—the two stages independently optimize for recall and precision, respectively, avoiding multi-objective conflicts inherent in a single model under asymmetric cost conditions. Second, module independence—each stage's model can be updated independently; when new fault types are added, only the second stage requires retraining, resulting in low system maintenance costs. Third, workflow alignment—the cascaded "screening → classification" logic is naturally consistent with the "alerting → diagnosis" standard workflow in aviation maintenance, facilitating integration into existing operational systems.

## 3.5 Knowledge Distillation-Based Precision–Recall Tuning Strategy

The architectural simplification from 3+1 branches to 1+1 branches inevitably incurs performance loss. Knowledge distillation achieves knowledge transfer by having the student model learn soft labels from the teacher's outputs. However, this paper discovers and exploits another core value of distillation: the systematic modulation of the precision–recall trade-off characteristics, enabling the same lightweight model to adapt to different deployment scenarios simply by switching the training approach.

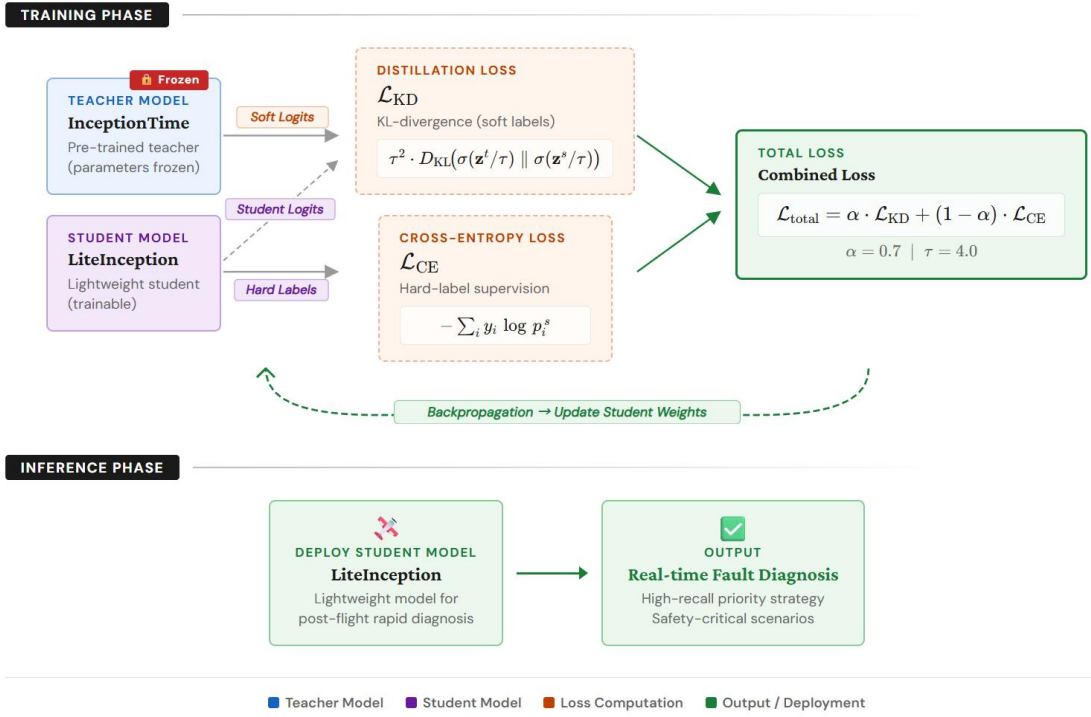

**Figure 4.** Distillation learning architecture diagram.

### 3.5.1 Distillation Loss Function

Knowledge distillation adopts a teacher–student architecture, where the teacher model is a pretrained InceptionTime (3+1 branches) and the student model is LiteInception (1+1 branches). During training, teacher parameters are frozen, and the student model's total loss function is:

$$L_{total} = \alpha \cdot L_{KD} + (1-\alpha) \cdot L_{CE} \tag{11}$$

where the hard-label loss $L_{CE} = -\sum_{i=1}^{K} y_i \log(p_i^s)$ is the standard cross-entropy, and the soft-label loss is:

$$L_{KD} = \tau^2 \cdot D_{KL}\left(\sigma\left(\frac{z^t}{\tau}\right) \| \sigma\left(\frac{z^s}{\tau}\right)\right) = \tau^2 \cdot \sum_{i=1}^{K} \sigma\left(\frac{z_i^t}{\tau}\right) \log \frac{\sigma(z_i^t / \tau)}{\sigma(z_i^s / \tau)} \tag{12}$$

where $z^t, z^s$ are the logits of the teacher and student models, respectively, $\sigma(\cdot)$ is the Softmax function, $\tau$ is the temperature coefficient, and $\alpha$ is the distillation weight. The $\tau^2$ factor balances gradient magnitudes at high temperatures.

### 3.5.2 Mechanism of Distillation for Precision–Recall Trade-Off Modulation

The mechanism by which knowledge distillation alters the model's P–R trade-off characteristics can be understood from three perspectives.

Decision boundary smoothing effect. Under hard-label training, the model's optimization objective is to push the probability of the correct class toward 1 and all others toward 0, tending to learn steep decision boundaries. For ambiguous samples near the boundary, the model lacks sufficient classification confidence and tends toward conservative predictions (i.e., "refusal to classify"), manifesting as high precision but low recall. Soft-label learning provides a smooth probability distribution as the supervisory signal, rendering the student model's decision boundaries more rounded. Ambiguous samples receive nonzero target probabilities in the soft labels, making the model more willing to render classification decisions on these samples, thereby improving recall at the cost of introducing more false positives.

Implicit encoding of inter-class relationships. The core role of the temperature coefficient $\tau$ is to control the degree of exposure of "dark knowledge" in the soft labels. When $\tau \to 1$, the Softmax output approaches a one-hot distribution, and inter-class relationship information is compressed. As $\tau$ increases, the output distribution tends toward uniformity, amplifying probability differences among non-target classes and exposing the teacher model's implicit knowledge of inter-class similarities—for example, the symptomatic overlap between "engine running rough" and "engine idle problems" would lead the teacher model to assign relatively similar probabilities to these two classes. After learning this structured information, the student model's discriminative ability for similar classes is enhanced, enabling more informative classification decisions when facing ambiguous samples.

Fundamental distinction from threshold adjustment. Conventional P–R tuning is achieved by shifting the classification threshold, which only changes the decision rule without altering the model's feature representation. Distillation training fundamentally changes the structure of the feature space learned by the model—soft labels guide the student model to learn a smoother representation that is more sensitive to inter-class relationships, which cannot be achieved through threshold adjustment alone.

**3.5.3 Scenario Adaptation Strategy**

Based on the above mechanism, the same LiteInception architecture acquires different P–R characteristics through two training approaches: direct training (hard labels only) tends toward high precision, suitable for assisted diagnosis scenarios with manual review; distillation training tends toward high recall, suitable for safety-critical scenarios where the cost of missed detections far exceeds that of false alarms. In general aviation aircraft fault diagnosis, the latter holds greater practical value—missing a potential fault may lead to a flight safety incident, whereas a false alarm merely increases re-inspection costs. The selection of distillation hyperparameters ($\tau$, $\alpha$) and their quantitative effects on the P–R trade-off are analyzed in detail in the sensitivity experiments in Section 4.4.

### 3.6 Dual-Level Interpretability Analysis Framework

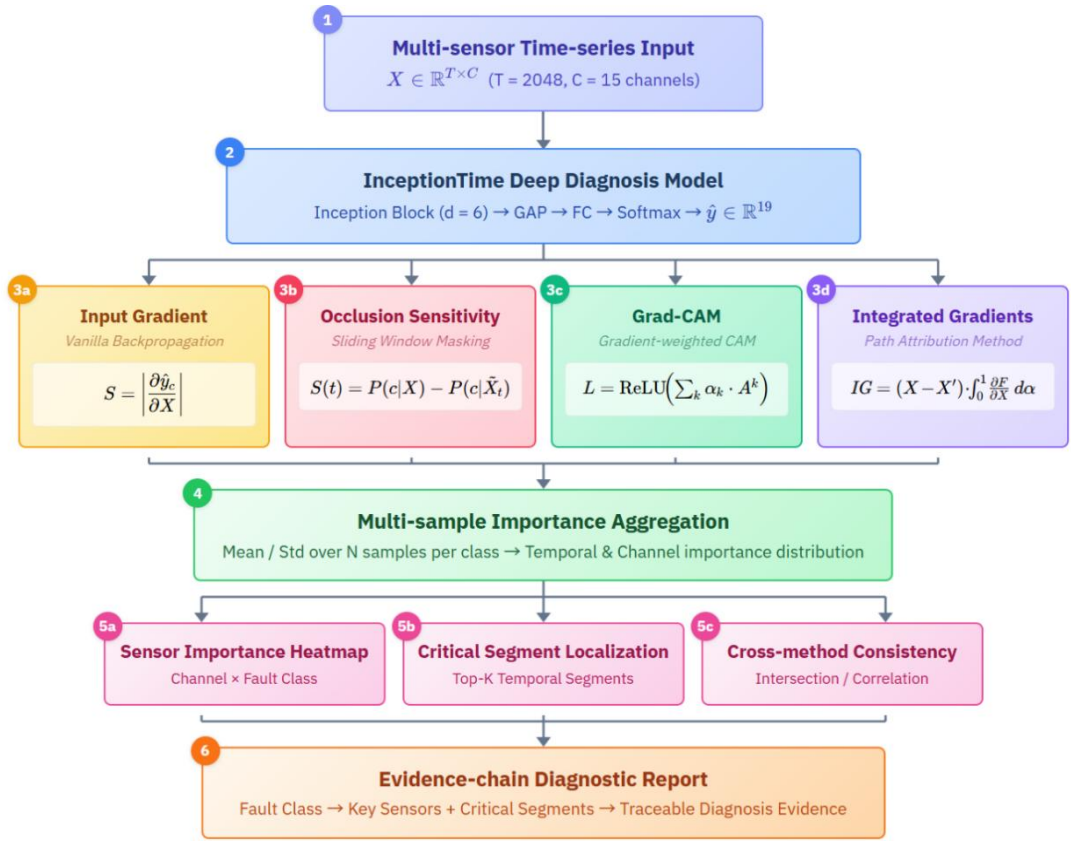

**Figure 5.** Interpretability architecture diagram.

The black-box nature of deep learning constrains its application in the aviation safety domain. Maintenance personnel need to know "which sensor" exhibited anomalies during "which time segment" to efficiently localize fault root causes. Existing interpretability research predominantly employs single attribution methods, lacking cross-validation to assess the robustness of interpretation results. This paper constructs a dual-level interpretability framework at the sensor level (channel dimension) and the temporal segment level (time dimension), integrating four complementary attribution methods for cross-validation.

### 3.6.1 Attribution Method Integration

**Table 1**

Comparison of Interpretability Methods

| Method | Core Idea | Computational Cost | Advantage | Limitation |
|---|---|---|---|---|
| Input Gradient | Absolute value of output partial derivative w.r.t. input $\partial \hat{y}_c / \partial X_{t,j}$ | Lowest (1 backward pass) | Computationally efficient | Gradient saturation |
| Occlusion Sensitivity | Change in target probability after sliding window occlusion $P(c\mid X) - P(c\mid \tilde{X})$ | Higher (multiple forward passes) | Gradient-free, intuitive and robust | Sensitive to window size |
| Grad-CAM | Gradient-weighted activation mapping of last-layer feature maps | Moderate | Smooth, stable, strong localization | Resolution limited by feature maps |
| Integrated Gradients | Gradient integral along the path from zero baseline to input, satisfying the completeness axiom | Highest ($m+1$ forward + backward passes) | Most theoretically rigorous, most accurate attribution | High computational overhead |

A comparison of the core characteristics of the four methods is shown in the table above.

(1) Input Gradient

This method computes the partial derivative of the model output with respect to the input via a single backward pass, using the absolute gradient value as the importance measure. Let the input sample be $\mathbf{X} \in \mathbb{R}^{T \times C}$ and the model's output for target class $c$ be $\hat{y}_c$. The importance score is:

$$S_{IG}(t, j) = \left| \frac{\partial \hat{y}_c}{\partial X_{t,j}} \right| \tag{13}$$

where $t$ denotes the time step and $j$ denotes the sensor channel. Summing along sensor channels yields temporal importance $S^{time}(t) = \sum_j S_{IG}(t, j)$, and summing along time yields channel importance $S^{ch}(j) = \sum_t S_{IG}(t, j)$. This method incurs the lowest computational overhead but may be affected by gradient saturation.

(2) Occlusion Sensitivity

This method sequentially occludes local regions of the input using a sliding window and observes the drop in target class probability. Let the window size be $w$ and the stride be $s$. With occlusion at position $[t_0, t_0 + w)$, the occluded input is denoted $\widetilde{\mathbf{X}}_{(t_0)}$, and the temporal importance is:

$$S_{OC}(t_0) = P(c \mid \mathbf{X}) - P(c \mid \widetilde{\mathbf{X}}_{(t_0)}) \tag{14}$$

Channel importance is obtained by zeroing out the entire $j$-th sensor channel and computing the probability difference. This method does not rely on gradient information and is intuitive and reliable, but requires multiple forward passes, resulting in higher computational cost.

(3) Grad-CAM

This method performs global average pooling on the gradients of the target class score with respect to the feature maps $\mathbf{A}^k \in \mathbb{R}^{T'}$ of the last Inception module, obtaining channel weights $\alpha_k$, and generates a class activation heatmap:

$$\alpha_k = \frac{1}{T'}\sum_t \frac{\partial \hat{y}_c}{\partial A_t^k}, \quad L_{\text{Grad-CAM}} = \text{ReLU}\left(\sum_k \alpha_k \cdot \mathbf{A}^k\right) \tag{15}$$

The resulting heatmap is upsampled to the original temporal resolution $T$ via bilinear interpolation, characterizing the contribution of different temporal segments to the classification decision. Compared to the Input Gradient method, Grad-CAM is smoother, more stable, and exhibits stronger localization capability.

(4) Integrated Gradients

This method accumulates gradients along the linear path from a zero baseline $\mathbf{X}'$ to the actual input $\mathbf{X}$, satisfying the completeness axiom, resulting in more accurate attribution. It is defined as:

$$\text{IG}_j(\mathbf{X}) = (X_j - X_j') \times \int_0^1 \frac{\partial F(\mathbf{X}' + \alpha(\mathbf{X} - \mathbf{X}'))}{\partial X_j} d\alpha \tag{16}$$

In implementation, the trapezoidal rule with $m = 50$ equally spaced interpolation points is used to approximate the integral. This method has the most rigorous theoretical foundation but the highest computational cost, requiring $m+1$ complete forward and backward passes.

The rationale for selecting these four methods lies in their complementarity: Input Gradient and Integrated Gradients are gradient-based methods, with the former being efficient but potentially unstable and the latter theoretically complete but computationally expensive; Grad-CAM leverages intermediate-layer feature map information, providing a perspective at a different granularity; Occlusion Sensitivity is entirely gradient-free, serving as an independent validation means.

**3.6.2 Dual-Level Attribution and Evidence Chain Generation**

For each fault category, $N = 30$ correctly classified samples are randomly selected, attribution scores are computed using all four methods, and the mean is taken to eliminate individual variation. Attribution results are aggregated along two dimensions:

Sensor-level attribution: The two-dimensional attribution matrix $S \in \mathbb{R}^{T \times C}$ from each method is summed along the temporal dimension to obtain a channel importance vector $S^{ch}(j) = \sum_{t} S(t, j)$. The top-5 sensor channel sets from the four methods are compared, and the intersection is taken as the high-confidence critical sensors.

Temporal segment-level attribution: The attribution matrix is summed along the sensor channel dimension to obtain a temporal importance curve $S^{time}(t) = \sum_{j} S(t, j)$. Critical temporal segment intervals are determined using the 90th percentile as the threshold.

Noise perturbation robustness validation. The multi-sample mean analysis above reveals statistical patterns at the category level but cannot verify whether attribution results for individual samples genuinely reflect the model's decision basis. To this end, this paper introduces noise perturbation contrastive analysis as a supplementary validation measure: Gaussian noise of varying intensities ($\sigma \in 0, 0.01, 0.03$) is applied to the same high-confidence sample, and the trend of attribution distribution changes as model confidence decreases is observed. The core logic is as follows: if the attribution method genuinely captures real features relevant to the model's decision, then when noise destroys these features, the attribution distribution should undergo corresponding structural changes—degenerating from a focused distribution on specific temporal segments and sensors to a diffuse distribution. Conversely, if the attribution results remain unchanged under noise perturbation, this suggests that they may reflect statistical biases in the data rather than the model's decision basis. Furthermore, in high-confidence samples, the spatial correspondence between waveform changes in the raw sensor signals (e.g., amplitude abrupt changes, trend inflection points) and the high-value regions in the temporal attribution provides intuitive corroboration of the physical plausibility of the attribution results.

Ultimately, by integrating the cross-validation results at the sensor level and the temporal segment level, a traceable evidence chain diagnostic report is generated at the granularity of "specific sensor × specific time segment," providing interpretable quantitative evidence for maintenance decisions.

## 4 Experiments

### 4.1 Dataset and Experimental Setup

#### 4.1.1 Dataset

This study uses the NGAFID (National General Aviation Flight Information Database) aviation maintenance dataset for experimental validation. The dataset originates from real flight records of Cessna 172 aircraft during normal operations, automatically collected by a flight school. The raw data contains 23 sensor channels at a sampling frequency of 1 Hz [8].

The data preprocessing pipeline is as follows: first, samples with more than 20% missing values and time series shorter than 2,000 time steps are removed; then, linear interpolation is applied to fill the remaining missing values (approximately 1% of data points contain NaN values); next, all time series are resampled to 2,048 time steps; finally, Min-Max normalization is applied to standardize each sensor channel to the [0, 1] range. The basic information of the preprocessed dataset is shown in the table below.

**Table 2**
Dataset overview

| Item | Value |
|---|---|
| Original sensor channels | 23 |
| Selected sensor channels | 15 |
| Time steps | 2,048 |
| Total samples | 11,446 |
| Post-maintenance samples (normal) | 5,844 |
| Pre-maintenance samples (faulty) | 5,602 |
| Number of fault categories | 19 |
| Train/test split | 80% / 20% |

Stage 1 (fault detection) divides the data into "normal" (post-maintenance) and "faulty" (pre-maintenance) classes. Stage 2 (fault identification) filters the 36 maintenance issue types to retain only those with ≥50 flights both before and after maintenance, resulting in 19 categories. The dataset is split into training and test sets using stratified sampling at an 8:2 ratio. To address the class imbalance in Stage 2 (largest class accounting for 18.33%, smallest class <1%), TimeWarp data augmentation is applied to minority classes with temporal axis warping (intensity 0.03), generating 2 augmented samples per original sample until all classes are balanced. The rationale for selecting TimeWarp is discussed in the ablation study in Section 4.5.

**4.1.2 Training Configuration**

All experiments are implemented using the PyTorch framework and run on an NVIDIA H100 GPU. The unified training configuration is as follows: Adam optimizer, learning rate $1\times10^{-4}$, weight decay $1\times10^{-4}$, batch size 32, 200 training epochs, ReduceLROnPlateau learning rate scheduler (factor=0.5, patience=10), and gradient clipping with max_norm=1.0.

**4.1.3 Baseline Methods and Evaluation Metrics**

**Table 3**
Baseline model configurations for fault detection (Stage 1)

| Model | Key hyperparameters |
|---|---|
| InceptionTime (3+1) | 6 Inception blocks, 3 conv branches (kernel size=3/5/7) + 1 MaxPool branch, nf=128 |
| CNN | 4 conv layers (channels 64→128→256→256, kernel size=7/5/5/3) |
| Transformer | d_model=128, heads=4, layers=2 |
| CNN-Transformer | CNN (channels 64→128) + Transformer (d_model=128, heads=4, layers=2) |
| LSTM | hidden_size=128, num_layers=2, bidirectional=True |
| ConvMHSA | Conv (channels 64→128) + MHSA (d_model=128, heads=4) |
| MiniRocket | num_kernels=84, kernel_sizes=[9,9,9] |
| LiteInception (1+1) | 6 Inception blocks, 1 conv branch (kernel size=3) + 1 MaxPool |

| Model | Key hyperparameters |
|---|---|
|  | branch, nf=64 |

To comprehensively evaluate the proposed method, the following baseline models are selected for comparison: InceptionTime (3+1) (our tuned version, serving as the teacher model), CNN (5-layer convolutional baseline), BiLSTM (bidirectional LSTM), Transformer (pure attention architecture), CNN-Transformer (hybrid architecture), ConvMHSA (lightweight attention hybrid), and MiniRocket.

**Table 4**
Baseline model configurations for fault identification (Stage 2)

| Model | Key hyperparameters |
|---|---|
| InceptionTime (3+1) | 6 Inception blocks, each with 3 conv branches (kernel size=3/5/7) + 1 MaxPool branch, nf=64, nb_filters=128, bottleneck=True, residual connections |
| CNN | 5 conv blocks (channels 64→128→256→256→512, kernel size=7/5/3/3/3), layer-wise MaxPool(2) downsampling, GAP, Dropout=0.5 |
| Transformer | Linear projection layer (15→128), sinusoidal positional encoding, 4-layer TransformerEncoder (d_model=128, 8-head attention, dim_ff=256, dropout=0.1), GAP |
| CNN-Transformer | 3-layer CNN (channels 64→128→128, kernel size=7/5/3, MaxPool(4) downsampling) + 2-layer TransformerEncoder (d_model=128, 8 heads, dim_ff=256, dropout=0.1), GAP |
| LSTM | 2-layer bidirectional LSTM (hidden_size=128, dropout=0.3), last time step output, Dropout=0.3 |
| ConvMHSA | 2-layer conv downsampling (64→128, kernel size=7/5, MaxPool(4)) + 2-layer MultiheadAttention (d_model=128, 8 heads) + LayerNorm residual, GAP |
| MiniRocket | 84 random fixed conv kernels (kernel size=7/9/11 alternating, max_dilation=32), only bias is trained; PPV features → 2-layer MLP (256, Dropout=0.5) |
| LiteInception (1+1) | 6 Inception blocks, each with 1 conv branch (kernel size=3) + 1 MaxPool branch, nf=64, nb_filters=128, bottleneck=True, residual connections |

It should be noted that the InceptionTime baseline used in this work is not the default configuration from the original paper [8], but has been systematically tuned based on the NGAFID dataset author's TensorFlow implementation for our specific tasks: kernel sizes were adjusted from [40, 20, 10] to [3, 5, 7] to adapt to the resampled 2,048-step sequences, nb_filters was increased from 32 to 128 to enhance fine-grained discriminability across 19 classes, and training strategies such as TimeWarp augmentation and learning rate scheduling were introduced. After tuning, accuracy improved from approximately 55% in the original version to approximately 80%. Using this fully optimized strong baseline as the teacher model and performance reference makes the evaluation of LiteInception's compression effectiveness more rigorous and credible.

**Table 5**
Hyperparameter tuning comparison

| Configuration | NGAFID original | Our tuned version |
|---|---|---|
| Framework | TensorFlow | PyTorch |
| Kernel sizes | [40, 20, 10] | [3, 5, 7] |
| Filters per branch (nb_filters) | 32 | 128 |
| Bottleneck dimension | 32 | 64 |
| Input shape | (4096, 23) | (2048, 15) |
| Batch size | 128 | 32 |
| Data augmentation | None | TimeWarp oversampling |
| Learning rate scheduling | None | ReduceLROnPlateau |
| Gradient clipping | None | max_norm=1.0 |

The evaluation metrics are Accuracy, Precision, Recall, and F1 score. Stage 1 prioritizes Recall as the primary metric (the cost of missed detections far exceeds that of false alarms), while Stage 2 prioritizes Macro F1 as the primary metric (ensuring equal evaluation of all classes in imbalanced data). Efficiency metrics include parameter count (Params), floating-point operations (FLOPs), GPU/CPU inference time, and peak memory usage.

## 4.2 Sensor Channel Selection Experiment

This section reports the experimental results of the multi-method fusion channel selection strategy described in Section 3.2.

## 4.2.1 Sensor Channel Importance Assessment Using Three Methods

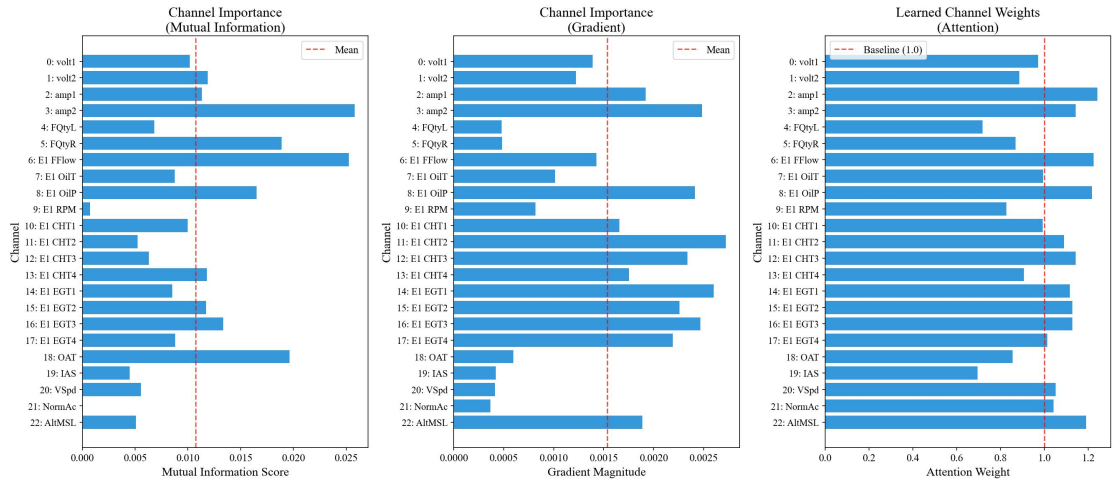

**Figure 6.** Feature importance analysis.

**Table 6**
Excluded sensor channels and rationale

| Sensor channel | Name | Reason for exclusion |
|---|---|---|
| 4, 5 | FQtyL, FQtyR | Ranked high by MI but low in both gradient and SE weights — a typical spurious correlation: fuel consumption correlates with flight duration, and certain faults are statistically associated with longer flights, but fuel quantity is not a direct fault indicator; moreover, channel 6 (E1 FFlow) already contains more diagnostically valuable dynamic information |
| 9 | E1 RPM | Actively regulated by the flight control system with minimal variation during normal operation; discriminative ability is insufficient across all three methods |
| 18 | OAT | An external environmental parameter with no direct causal relationship to internal engine faults; the high MI may originate from seasonal confounding factors |
| 19–22 | IAS, VSpd, AltMSL, NormAc | Reflect flight mission characteristics rather than engine health status; causal direction error — these are consequences of faults rather than causes |

The figure shows the importance scores of all 23 original sensor channels evaluated by three methods: mutual information, gradient analysis, and SE attention weights. The three methods exhibit partially consistent and partially divergent assessment results, validating the necessity of multi-method fusion.

Following the multi-method cross-validation strategy described in Section 3.2, and considering the ranking consistency across three methods combined with domain-specific physical knowledge, 8 sensor channels are ultimately excluded, retaining 15 sensor channels (indices: 0, 1, 2, 3, 6, 7, 8, 10, 11, 12, 13, 14, 15, 16, 17). The excluded channels and their rationale are shown in the table above.

The retained 15 sensor channels cover five core functional subsystems: electrical system (channels 0–3: volt1, volt2, amp1, amp2 — mapping the electrical signatures of engine mechanical loads), fuel flow (channel 6: E1 FFlow — monitoring combustion process dynamics), lubrication system (channels 7–8: E1 OilT, E1 OilP — oil temperature and pressure as early degradation indicators), cylinder head temperature (channels 10–13: E1 CHT1–CHT4 — reflecting the thermal balance of each cylinder), and exhaust gas temperature (channels 14–17: E1 EGT1–EGT4 — directly measuring combustion efficiency).

**4.2.2 Validation of Channel Selection Effectiveness**

It should be noted that sensor channel selection, as a pre-analysis step in the data preprocessing stage, aims to evaluate the diagnostic contribution of sensor channels and reduce input signal-to-noise ratio using the existing strong baseline model (InceptionTime 3+1). This stage does not involve data augmentation or other training optimization strategies (data augmentation will be introduced during the formal training of LiteInception).

To validate the effectiveness of the channel selection strategy, this study compares the performance of InceptionTime (3+1) on the Stage 2 fault identification task using all 23 sensor channels versus the selected 15 sensor channels. Both experimental groups use the same model architecture, training configuration, and data split, with no data augmentation applied — only the number of input sensor channels differs.

Through the above multi-method cross-validation and physical causality analysis, the input dimensionality is reduced from 23 to 15 sensor channels, removing spurious correlation channels (e.g., fuel quantity remaining), channels with insufficient discriminative power (e.g., E1 RPM), and channels with incorrect causal direction (e.g., flight state parameters), while retaining core sensors from five subsystems — electrical, fuel, lubrication, temperature, and exhaust — that are directly related to engine fault mechanisms. Experimental results demonstrate that model performance actually improves after selection to 15 channels, while the parameter count is also reduced. This validates that the excluded 8 sensor channels indeed contain redundant or interfering information — the removal of spuriously correlated channels (e.g., fuel quantity remaining) and causally misaligned channels (e.g., flight state parameters) allows the model to focus on signals directly related to fault mechanisms, effectively reducing the input signal-to-noise ratio and learning more discriminative feature representations. All subsequent formal experiments are conducted using the selected 15 sensor channels.

Table 7

Sensor channel selection ablation study
(Stage 2, InceptionTime 3+1, No Data Augmentation)

| Input sensor channels | Acc | Macro F1 | Precision | Recall |
|---|---|---|---|---|
| 23 channels (all) | 69.69% | 69.58% | 63.12% | 65.66% |
| **15 channels (selected)** | **77.58%** | **78.23%** | **75.92%** | **76.86%** |

## 4.3 Fault Detection Experiment (Stage 1)

This section evaluates the performance of the first-stage fault detection model.

Table 8

Fault detection results comparison

| Model | Params | FLOPs | Acc | F1 | Precision | Recall | GPU time |
|---|---|---|---|---|---|---|---|
| **LiteInception (1+1) + Transformer** | 933K | 3.82G | **81.92%** | **81.84%** | 80.49% | 83.24% | 1.76ms |
| InceptionTime (3+1) | 2.15M | 8.81G | 79.82% | 80.68% | 75.92% | 86.06% | 1.25ms |

| Model | Params | FLOPs | Acc | F1 | Precision | Recall | GPU time |
|---|---|---|---|---|---|---|---|
| ConvMHSA | 181K | 742M | 79.77% | 80.67% | 75.77% | 86.26% | 0.58ms |
| CNN-Transformer | 331K | 1.35G | 77.91% | 79.52% | 72.79% | 87.62% | 0.90ms |
| CNN | 411K | 1.68G | 78.58% | 78.89% | 76.20% | 81.77% | 0.31ms |
| MiniRocket | 108K | 443M | 69.75% | 69.04% | 69.18% | 68.91% | 0.62ms |
| LSTM | 578K | 2.37G | 60.59% | 66.61% | 56.91% | 80.31% | 13.27ms |
| Transformer | 268K | 1.10G | 51.62% | 66.56% | 50.30% | 98.34% | 0.69ms |

The proposed LiteInception+Transformer hybrid model achieves the best F1 score (81.84%) while delivering significant efficiency improvements: compared to InceptionTime, the parameter count is reduced by 57% (2.15M → 933K), FLOPs are reduced by 57% (8.81G → 3.82G), while F1 actually improves by 1.16 percentage points.

In terms of the Precision-Recall trade-off, the proposed model exhibits a more balanced performance (Precision 80.49%, Recall 83.24%, a difference of only 2.75pp), whereas InceptionTime shows a P-R gap of 10.14pp (Precision 75.92% vs. Recall 86.06%), indicating a higher rate of false positives. Compared to other baselines, the proposed model leads comprehensively at equal or fewer parameters: F1 improves by 1.17pp over ConvMHSA (181K parameters) and by 2.32pp over CNN-Transformer (331K parameters), with Precision substantially increasing by 7.70pp.

The pure Transformer and LSTM perform poorly on time series classification (F1 ≈ 66%), corroborating the analysis in Section 3.3.2: for aviation sensor signals with distinct local pattern characteristics, the inductive bias of convolution is superior to pure attention mechanisms. The LiteInception+Transformer hybrid architecture successfully combines the strengths of both — convolution for extracting local patterns and attention for capturing long-range dependencies.

It is worth noting that as the safety gateway of the cascaded system, the Stage 1 Recall (83.24%) directly determines whether potentially faulty flights can enter subsequent fine-grained diagnosis — undetected anomalies will completely escape the entire diagnostic pipeline. The current Recall of 83.24% means approximately 16.76% of truly anomalous flights are misclassified as normal. For high-safety deployment scenarios, the binary classification threshold can be adjusted to further improve Recall, at the cost of some Precision degradation.

### 4.4 Fault Identification and Knowledge Distillation Experiment (Stage 2)

#### 4.4.1 Multi-Model Comparison Experiment

Table 9

Fault Identification Results Comparison

| Model | Params | FLOPs | Acc | Macro F1 | GPU time | CPU time | Peak memory |
|---|---|---|---|---|---|---|---|
| **LiteInception (1+1)** | 639K | 2.58G | 77.00% | 75.41% | 1.02ms | 6.35ms | 89MB |
| InceptionTime (3+1) | 2,159K | 8.77G | 80.41% | 79.18% | 1.31ms | 52.37ms | 128MB |
| ConvMHSA | 183K | 0.07G | 47.37% | 41.98% | 0.48ms | 1.73ms | 121MB |
| CNN-Transformer | 365K | 0.08G | 47.76% | 39.90% | 0.73ms | 2.33ms | 110MB |
| CNN | 748K | 0.41G | 74.76% | 73.16% | 0.43ms | 1.80ms | 101MB |
| MiniRocket | 71K | 0.09G | 6.82% | 4.80% | 3.49ms | 30.68ms | 119MB |
| LSTM | 549K | 0.60G | 4.00% | 4.55% | 13.01ms | 192.21ms | 186MB |
| Transformer | 534K | <0.01G | 8.97% | 10.46% | 1.41ms | 42.45ms | 112MB |

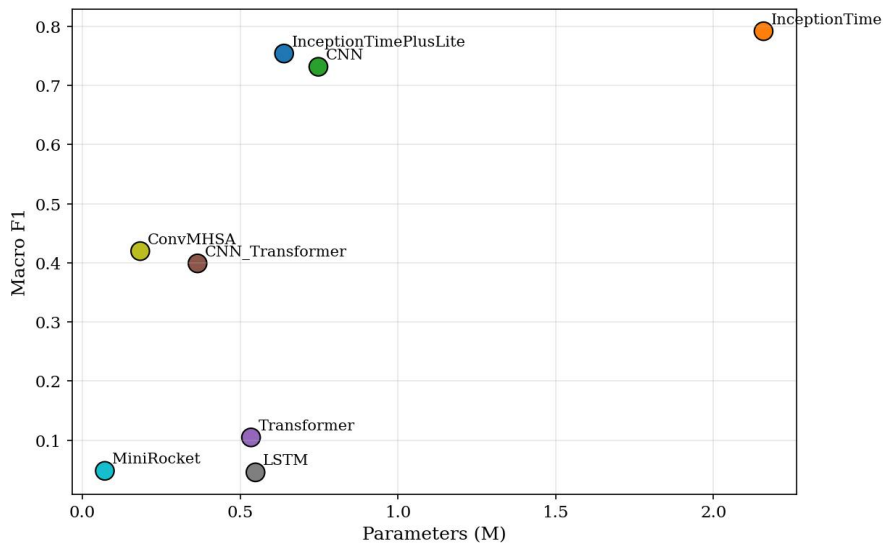

**Figure 7.** Accuracy-efficiency scatter plot.

This section evaluates the performance on the 19-class fault identification task in Stage 2. This task features severely imbalanced sample distributions across classes, imposing higher demands on model discriminability and generalization capability.

The experimental results exhibit a clear performance stratification. The Inception family (InceptionTime and LiteInception) forms the first tier, validating the superiority of the multi-scale parallel convolution structure for long-sequence multi-channel sensor data. The plain CNN ranks third with 74.76% accuracy, confirming the natural adaptability of convolution operations to time series signals. Hybrid architectures containing Transformer components (CNN-Transformer, ConvMHSA) lag significantly behind with accuracy below 50%, indicating that self-attention mechanisms introduce excessive noisy interactions at the current data scale. The pure Transformer, MiniRocket, and BiLSTM almost completely fail on the 19-class task (F1 < 11%) and have no practical applicability.

Compression effectiveness analysis of LiteInception. Compared to the tuned InceptionTime, LiteInception achieves a 70.4% parameter reduction (2,159K → 639K), 70.6% FLOPs reduction (8.77G → 2.58G), 8.2× CPU inference speedup (52.37ms → 6.35ms), and 30% memory reduction (128MB → 89MB). In terms of performance, Accuracy decreases by 3.4pp (80.41% → 77.00%) and Macro F1 decreases by 3.8pp (79.18% → 75.41%), but it still significantly surpasses the third-ranked CNN (F1 higher by 2.3pp). Core conclusion: trading approximately 3–4pp of F1 loss for 70% parameter compression and 8× CPU speedup demonstrates significant engineering advantages in the efficiency-accuracy trade-off.

**4.4.2 Knowledge Distillation Experiment**

The distillation experiment is conducted independently from the multi-model comparison above to validate the Precision-Recall adjustment mechanism described in Section 3.5. The teacher model is the tuned InceptionTime (trained once), and the student model is compared under direct training (1 run) and distillation training (repeated 10 runs). Distillation configuration: $\tau = 8.0$, $\alpha = 0.7$.

The experimental results validate the theoretical analysis in Section 3.5.2 — distillation induces a significant rebalancing between Precision and Recall. The directly trained student model exhibits a high-Precision, low-Recall characteristic (77.92% vs. 73.46%, a gap of 4.46pp), indicating conservative prediction tendencies; after distillation, Precision decreases by approximately 1.5pp, but Recall increases substantially by approximately 2.4pp (73.46% → 75.85%), narrowing the P-R gap from 4.46pp to 0.55pp for a more balanced distribution.

This phenomenon is consistent with the decision boundary smoothing effect of soft labels: the smoothed probability distribution from the teacher output at high temperature $\tau$ makes the student model more inclined to make classification decisions for ambiguous samples, thereby improving recall. In terms of overall metrics, the best distillation F1 (75.67%) slightly outperforms direct training (75.18%) by 0.49pp; the low standard deviation across 10 experiments (F1 std = 0.42%) indicates good stability of the distillation process.

**Table 10**
Knowledge distillation results

| Model | Params | Acc | Macro F1 | Precision | Recall |
|---|---|---|---|---|---|
| Teacher: InceptionTime (3+1) | 2,159K | 79.82% | 77.75% | 80.05% | 76.17% |
| LiteInception (direct training) | 639K | 77.00% | 75.18% | 77.92% | 73.46% |
| LiteInception (distillation, best) | 639K | 77.68% | 75.67% | 76.40% | 75.85% |
| LiteInception (distillation, mean±std) | 639K | 76.96±0.48% | 75.03±0.42% | 75.20±0.77% | 75.76±0.46% |

**Table 11**
Scenario-Specific recommendations for the two training strategies

| Training strategy | Advantage | Recommended scenario |
|---|---|---|
| Direct training | High Precision (77.92%), fewer false positives | Assisted diagnosis with human review |
| Knowledge | High Recall (75.85%), fewer | Safety-critical automated |

| Training strategy | Advantage | Recommended scenario |
|---|---|---|
| distillation | missed detections | alerting |

## 4.5 Ablation and Sensitivity Analysis

To validate the effectiveness of each core design decision, this section conducts ablation and sensitivity experiments across four key dimensions. All experiments are performed on the Stage 2 task using the same data split (training set: 4,102 samples; test set: 1,026 samples).

### 4.5.1 Branch Number Ablation

The multi-branch parallel design is the core feature of InceptionTime. To validate the rationality of the 1+1 branch simplification, four configurations are compared: 1+0, 1+1, 2+1, and 3+1.

Table 12
Branch number ablation study

| Architecture | Params | Relative Params | Acc | F1 | F1 Loss |
|---|---|---|---|---|---|
| 3+1 | 2.159M | 100% | 80.12% | 78.47% | Baseline |
| 2+1 | 1.301M | 60.3% | 79.43% | 77.67% | −0.80pp |
| **1+1** | 0.639M | 29.6% | 77.68% | 75.72% | −2.75pp |
| 1+0 | 0.303M | 14.0% | 74.85% | 73.46% | −5.01pp |

The results support the theoretical analysis in Section 3.3.2. From 3+1 to 2+1, F1 drops by only 0.80pp, indicating that the third convolutional branch ( $k = 7$ ) has limited marginal contribution. The 1+1 architecture maintains approximately 96.5% of the F1 score with less than 30% of the baseline's parameters. Removing the MaxPool branch (1+0) causes a sharp F1 drop of 5.01pp, validating the irreplaceable role of MaxPool in capturing temporal extremal features.

### 4.5.2 Data Augmentation Strategy Ablation

To address the class imbalance in the 19-class fine-grained classification task, six augmentation strategies are compared against a no-augmentation baseline.

The results exhibit a stark polarization: temporal-domain deformation strategies (TimeWarp, Smooth, Window Slice) produce positive gains, while amplitude-domain strategies (Gaussian Noise, Magnitude Shift/Scale) cause performance degradation. TimeWarp leads with a +13.06% improvement, as its nonlinear temporal axis warping simulates real flight variations such as RPM fluctuations and sampling jitter, closely matching actual operational scenarios. The failure of amplitude-domain strategies is attributable to the fact that normalized sensor amplitudes carry important physical semantics (absolute temperature values, pressure ratios, etc.), and random amplitude perturbation destroys these physical constraints.

Table 13

Data augmentation strategy ablation

| Augmentation strategy | F1 | Relative to baseline |
|---|---|---|
| No augmentation (baseline) | 67.69% | — |
| **TimeWarp (temporal warping)** | **76.53%** | **+13.06%** |
| Smooth (smoothing) | 75.07% | +10.90% |
| Window Slice (window slicing) | 73.03% | +7.88% |
| Gaussian Noise | 64.51% | −4.69% |
| Magnitude Shift | 51.91% | −23.32% |
| Magnitude Scale | 41.72% | −38.37% |

**4.5.3 Distillation Hyperparameter Sensitivity**

A $4\times 4$ grid search ( $\tau \in 2,4,6,8$, $\alpha \in 0.3, 0.5, 0.7, 0.9$ ) explores the distillation parameter space, totaling 16 experimental configurations.

Table14

Top-5 distillation hyperparameter search results

| Rank | τ | α | Acc | Precision | Recall | F1 |
|---|---|---|---|---|---|---|
| 1 | 8.0 | 0.70 | 78.36% | 76.91% | 76.81% | 76.37% |
| 2 | 6.0 | 0.90 | 77.49% | 76.55% | 75.98% | 75.78% |
| 3 | 2.0 | 0.30 | 77.00% | 78.61% | 73.61% | 75.42% |
| 4 | 8.0 | 0.30 | 76.41% | 75.83% | 75.65% | 75.35% |
| 5 | 6.0 | 0.70 | 76.61% | 76.16% | 75.19% | 75.27% |

High temperature ($\tau = 6 \sim 8$) combined with medium-to-high weight ($\alpha = 0.7 \sim 0.9$) yields the best performance, consistent with the theoretical expectation in Section 3.5.2: high temperature sufficiently softens the probability distribution to expose inter-class relationships. The F1 range across all 16 experiments is 73.75%–76.37%, with a maximum fluctuation of only 2.62pp, demonstrating good robustness of the distillation framework to hyperparameter choices. Notably, the third-ranked low-temperature, low-weight combination ($\tau = 2$, $\alpha = 0.3$) exhibits the highest Precision (78.61%) and the lowest Recall (73.61%), further corroborating the regulatory role of the temperature parameter on the Precision-Recall trade-off.

**4.5.4 Network Depth and Kernel Size Sensitivity**

**Table 15**

Network depth sensitivity (1+1 architecture)

| Depth | Acc | F1 | Precision | Recall |
|---|---|---|---|---|
| 3 | 66.96% | 67.81% | 68.51% | 67.88% |
| 6 | 78.85% | 76.31% | 79.17% | 74.08% |
| 9 | 78.46% | 78.00% | 81.26% | 75.61% |
| 12 | 77.58% | 76.64% | 79.29% | 74.78% |

**Table 16**

Kernel size sensitivity (1+1 architecture, depth=6)

| Kernel size | Acc | F1 | Precision | Recall |
|---|---|---|---|---|
| 3 | 78.85% | 76.31% | 79.17% | 74.08% |
| 9 | 78.27% | 75.86% | 78.28% | 74.52% |
| 21 | 77.19% | 74.46% | 77.59% | 72.60% |
| 39 | 77.68% | 75.84% | 78.78% | 73.87% |

F1 jumps by 8.5pp from depth 3 to 6, representing the most significant gain interval; depth 9 achieves the optimal F1 (78.00%), but depth 12 shows a decline (76.64%), potentially due to overfitting. Considering overall efficiency, depth 6 is selected as the default configuration. Regarding kernel size, $k = 3$ achieves the best F1, consistent with the receptive field analysis in Section 3.3.2: small kernels combined with deep stacking cover the required receptive field while preserving fine temporal resolution, outperforming the single-layer large-kernel strategy.

## 4.6 Interpretability Analysis

This section applies the dual-layer interpretability framework constructed in Section 3.6 to analyze the model's decision-making mechanism. For each fault category, 30 correctly classified test samples are randomly selected, and four attribution methods are applied respectively, with results averaged.

### 4.6.1 Sensor-Level Attribution Analysis

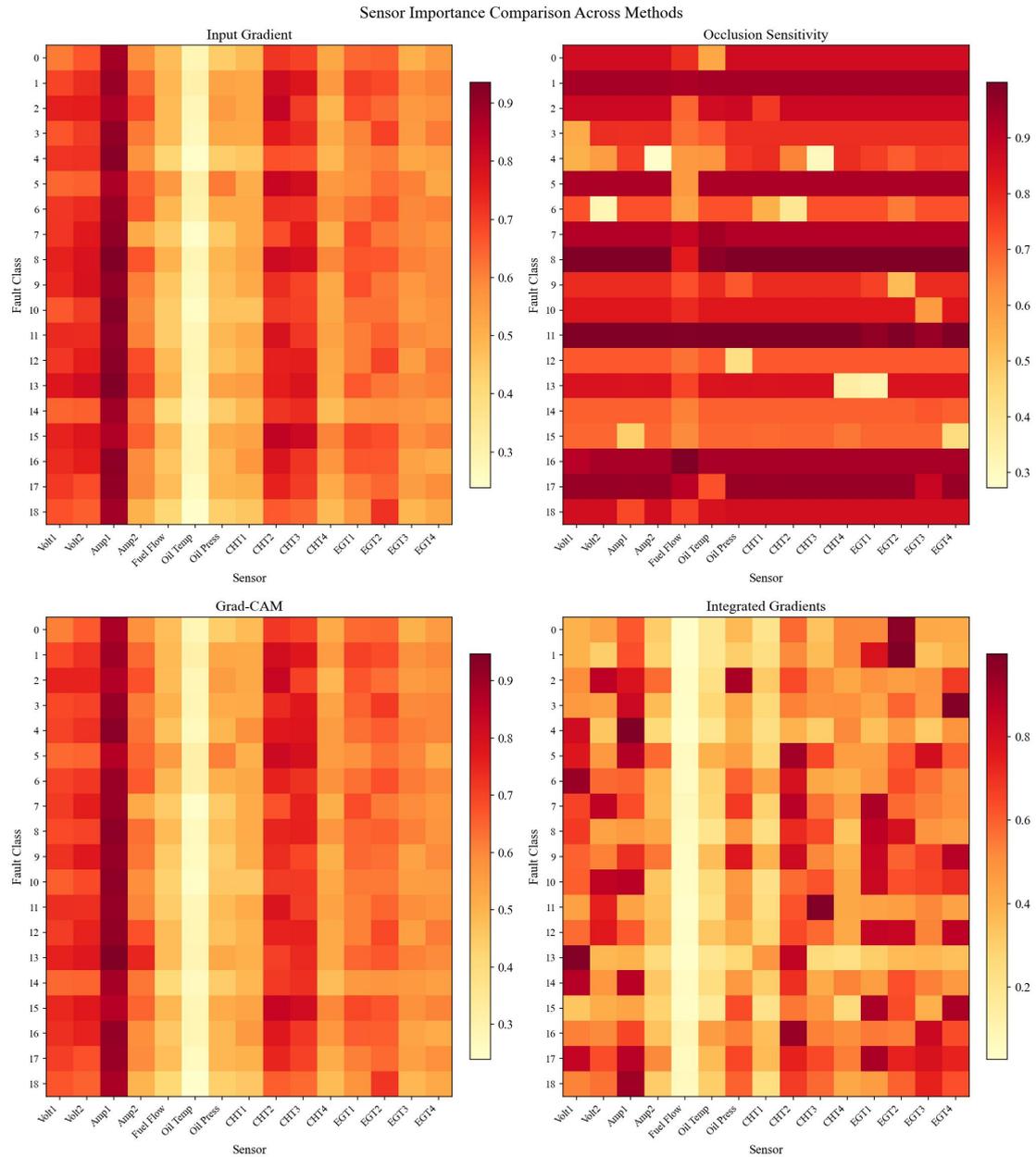

**Figure 8.** Sensor importance heatmap (comparison of four methods).

The following key findings can be extracted from the sensor-level attribution:

The central role of primary battery current (amp1). Across all 19 fault categories, the primary battery current sensor consistently exhibits extremely high importance (average score above 0.90 in Input Gradient and Grad-CAM methods), ranking as Top-1 in faults such as rocker cover leak/loose/damage (Category 4, Integrated Gradients score 0.993). This indicates that the current signal is a core indicator of engine health status. This is consistent with the physical characteristics of piston engines — current directly reflects engine load and operating state, and any fault causing load changes will be manifested in the current signal.

Differentiated response of cylinder head temperatures. E1 CHT2 and E1 CHT3 show higher importance in most fault categories, but different faults exhibit differentiated patterns: in oil cooler maintenance faults (Category 5), E1 CHT2 serves as the Top-1 sensor at 0.927; in baffle tie/tie rod loose or damage (Category 11), E1 CHT3 ranks Top-1 at 0.988 while E1 CHT2 ranks Top-3 at 0.625; in baffle crack/damage/loose/miss (Category 8), E1 CHT2 ranks Top-3 (0.718). Overall, E1 CHT2 contributes more prominently in core engine faults and oil system faults, while E1 CHT3 shows relatively higher importance in baffle structural faults (Categories 7–8, 10–11), reflecting the differential impact of different faults on cylinder thermal distribution.

Specificity-driven contribution of exhaust gas temperatures. The EGT series shows significant value in specific faults: in engine failure/fire/time out (Category 15), both E1 EGT1 and E1 EGT4 have importance scores of 0.910; in engine idle/RPM issue (Category 12), three EGT sensors — E1 EGT4 (0.871), E1 EGT1 (0.855), and E1 EGT2 (0.843) — are simultaneously activated; in engine seal/tube/bolt loose or damage (Category 1), E1 EGT2 reaches 0.998; in baffle crack/damage/loose/miss (Category 8), E1 EGT1 (0.872) and E1 EGT2 (0.797) simultaneously rank in the Top-2. This is consistent with the sensitivity of exhaust gas temperature to combustion anomalies and changes in gas tightness.

Auxiliary diagnostic value of voltage and oil pressure. In cylinder crack/fail/need part repair (Category 2), E1 OilP ranks Top-1 (0.913); in engine failure/fire/time out (Category 15), E1 OilP ranks Top-3 (0.640), reflecting the sensitivity of the oil system to engine mechanical integrity faults. volt2 contributes significantly in cylinder crack/fail/need part repair (Category 2, 0.873) and baffle tie/tie rod loose or damage (Category 11, 0.742). volt1 ranks Top-2 or Top-3 in rocker cover leak/loose/damage (Category 4, 0.815) and cylinder compression issue (Category 17, 0.851), indicating that voltage signals have auxiliary diagnostic value for specific structural faults.

Table 17
Key sensors and importance scores for representative fault categories (integrated gradients)

| Fault category | Fault came | Top-1 sensor | Score | Top-2 sensor | Score | Top-3 sensor | Score |
|---|---|---|---|---|---|---|---|
| 0 | Engine run rough | E1 EGT2 | 0.969 | amp1 | 0.611 | E1 CHT2 | 0.577 |
| 1 | Engine seal/tube/bolt loose or damage | E1 EGT2 | 0.998 | E1 EGT1 | 0.782 | amp1 | 0.632 |
| 2 | Cylinder crack/fail/need part repair | E1 OilP | 0.913 | volt2 | 0.873 | amp1 | 0.785 |
| 4 | Rocker cover leak/loose/damage | amp1 | 0.993 | volt1 | 0.815 | E1 CHT4 | 0.519 |
| 5 | Oil cooler need maintenance | E1 CHT2 | 0.927 | amp1 | 0.889 | E1 EGT3 | 0.808 |
| 8 | Baffle crack/damage/loose/miss | E1 EGT1 | 0.872 | E1 EGT2 | 0.797 | E1 CHT2 | 0.718 |
| 11 | Baffle tie/tie rod loose or damage | E1 CHT3 | 0.988 | volt2 | 0.742 | E1 CHT2 | 0.625 |
| 12 | Engine idle/RPM issue | E1 EGT4 | 0.871 | E1 EGT1 | 0.855 | E1 EGT2 | 0.843 |
| 15 | Engine failure/fire/time out | E1 EGT1 | 0.910 | E1 EGT4 | 0.910 | E1 OilP | 0.640 |
| 16 | Engine/propeller overspeed or damage | E1 CHT2 | 0.948 | E1 EGT3 | 0.837 | amp1 | 0.655 |
| 13 | Intake gasket leak/damage | volt1 | 0.989 | E1 CHT2 | 0.860 | E1 CHT1 | 0.480 |
| 17 | Cylinder compression issue | E1 EGT1 | 0.913 | amp1 | 0.881 | volt1 | 0.851 |

## 4.6.2 Temporal Segment Attribution Analysis

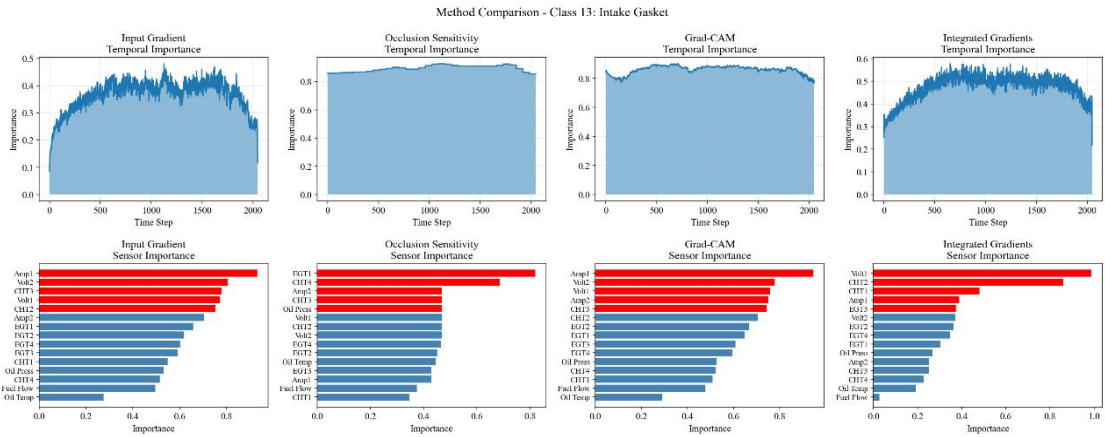

**Figure 9.** Multi-sample mean temporal dimension analysis.

**Table 18**

Key temporal segments for representative fault categories

| Fault category | Fault name | Input gradient | Occlusion sensitivity | Grad-CAM | Integrated gradients |
|---|---|---|---|---|---|
| 0 | Engine run rough | 48%–90% | 28%–56% | 34%–82% | 18%–88% |
| 1 | Engine seal/tube/bolt loose or damage | 23%–85% | 0%–100% | 0%–92% | 14%–93% |
| 2 | Cylinder crack/fail/need part repair | 41%–90% | 12%–22% | 0%–66% | 16%–90% |
| 5 | Oil cooler need maintenance | 30%–89% | 59%–84% | 0%–98% | 20%–91% |
| 8 | Baffle crack/damage/loose/miss | 23%–90% | 78%–87% | 0%–93% | 20%–90% |
| 12 | Engine idle/RPM issue | 29%–91% | 59%–87% | 19%–68% | 5%–99% |
| 13 | Intake gasket leak/damage | 27%–90% | 50%–87% | 22%–63% | 25%–87% |
| 15 | Engine failure/fire/time out | 29%–86% | 62%–84% | 33%–74% | 25%–86% |
| 16 | Engine/propeller overspeed or damage | 20%–89% | 0%–12% | 8%–90% | 14%–91% |
| 17 | Cylinder compression issue | 30%–90% | 0%–9% | 33%–100% | 22%–90% |

The key diagnostic information for most faults is concentrated in the mid-to-late segments of the flight record (30%–90% interval), likely related to the cumulative effect of faults and performance degradation under sustained engine operation. The granularity of key segment identification varies across methods: occlusion sensitivity, due to its fixed window approach, provides more concentrated and precise localization (e.g., 78%–87% for Category 8 baffle crack and 62%–84% for Category 15 engine failure), while integrated gradients, by considering the complete path, yields broader ranges (e.g., 5%–99% for Category 12).

Notably, occlusion sensitivity analysis for certain fault categories reveals key segments concentrated in the early flight phase: cylinder crack/fail/need part repair (Category 2) at 12%–22%, cylinder compression issue (Category 17) at 0%–9%, and engine/propeller overspeed or damage (Category 16) at 0%–12%. This finding has significant engineering implications — it indicates that these mechanical structural faults leave identifiable feature signatures in sensor data during the early flight phase, providing a basis for early fault warning. Meanwhile, engine seal/tube/bolt loose or damage (Category 1) exhibits full-range importance (0%–100%) in occlusion sensitivity, suggesting that the signal characteristics of such faults are distributed throughout the entire flight cycle, reflecting the physical nature of their gradual leakage behavior.

Single-sample temporal segment attribution verification under noise perturbation. The above analysis is based on the mean of 30 samples, reflecting category-level statistical patterns. To further verify the authenticity of temporal segment attribution on individual samples, this study applies three levels of Gaussian noise ($\sigma = 0, 0.01, 0.03$) to the highest-confidence correctly classified sample for each category and observes the changes in attribution distribution.

Taking Category 13 (intake gasket leak/damage) as an example (as shown in the figure), the original sample confidence is 0.854, which drops to 0.626 after mild noise and further to 0.339 after heavy noise. The attribution distribution exhibits the following systematic changes:

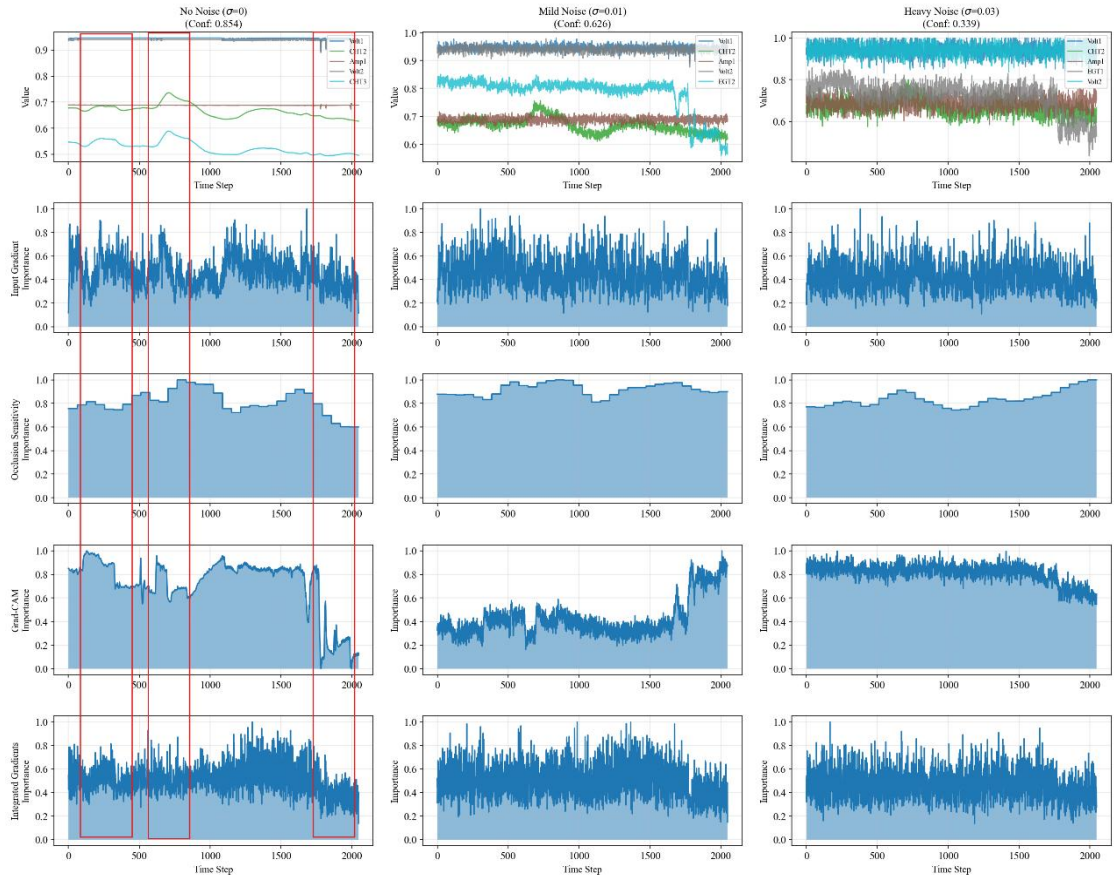

**Figure 10.** Single-sample multi-method temporal dimension analysis (noise perturbation comparison).

Under the no-noise condition ($\sigma = 0$), the temporal attribution from all four methods presents clear structured patterns, and an observable spatial correspondence exists between the high-attribution regions and the variation characteristics of the raw sensor waveforms: Grad-CAM assigns high attribution values at the peaks or troughs of the raw sensor signals, coinciding with the significant fluctuation intervals of E1 CHT2 and E1 CHT3 signals at those time segments; occlusion sensitivity exhibits step-like high-value regions at raw sensor peaks or troughs. This correspondence indicates that the model assigns higher decision weights to temporal segments where signals undergo significant changes, consistent with the physical expectation that fault signatures typically accompany abnormal sensor reading variations.

As noise intensity increases, the attribution distributions from all four methods degrade from structured patterns to diffuse distributions. Input gradient attribution becomes nearly uniformly distributed across all time steps under heavy noise, losing discriminative ability; Grad-CAM's attribution peak regions spread from the initial concentrated area to the full time span; the step structure of occlusion sensitivity tends toward flatness. This degradation trend is highly synchronized with the confidence decline (0.854 → 0.339), validating from a counterfactual perspective the causal relationship between attribution results and the model's decision mechanism — once noise destroys the features that the model actually relies upon, the attribution distribution becomes correspondingly diffuse.

The four methods exhibit different degrees of sensitivity to noise perturbation: occlusion sensitivity maintains a relatively clear step structure under mild noise, demonstrating strong robustness, consistent with its mechanism of directly measuring output changes without relying on gradients; input gradient is the most noise-sensitive method, with mild noise causing significant changes in the attribution pattern, reflecting the high sensitivity of gradients to small input perturbations. Cross-category noise perturbation experiments show consistent patterns: the degree of attribution structuredness for high-confidence samples is generally higher than for low-confidence samples, and noise consistently causes diffuse degradation of attributions, further supporting the necessity of multi-method ensembling.

**4.6.3 Multi-Method Consistency Verification**

Note: For Category 3 (intake tube/bolt/seal/boot loose or damage) and Category 13 (intake gasket leak/damage), the intersection of the Top-5 sensor sets across all four methods is empty, indicating that these two fault types exhibit significant divergence in signal patterns under different attribution perspectives.

Multi-method cross-validation reveals a clear hierarchical structure in method consistency across different fault categories. Oil cooler need maintenance (Category 5) exhibits the highest method consistency, with all four methods identifying E1 CHT2, E1 CHT3, and amp1 as key sensors, indicating that this fault has a clear and stable physical signature pattern. Baffle seal loose/damage (Category 6) and baffle tie/tie rod loose or damage (Category 11) also share three consensus sensors, suggesting that baffle structural faults have relatively well-defined sensor response patterns. In contrast, cylinder crack/fail/need part repair (Category 2) has only one consensus sensor (amp1), reflecting the more complex signal pattern of this fault, with different methods capturing its features from different perspectives.

Table 19

Key sensors unanimously identified by all four methods

| Fault category | Fault name | Unanimously identified key sensors |
|---|---|---|
| 0 | Engine run rough | E1 CHT2, amp1 |
| 1 | Engine seal/tube/bolt loose or damage | E1 CHT2, amp1 |
| 2 | Cylinder crack/fail/need part repair | amp1 |
| 4 | Rocker cover leak/loose/damage | volt1, amp1 |
| 5 | Oil cooler need maintenance | E1 CHT2, E1 CHT3, amp1 |
| 6 | Baffle seal loose/damage | volt1, E1 CHT2, amp1 |
| 7 | Baffle rivet loose/miss/damage | volt2 |
| 8 | Baffle crack/damage/loose/miss | volt1, E1 CHT3 |
| 9 | Baffle plug need repair/replace | E1 CHT2, amp1 |
| 10 | Baffle screw miss/loose | volt2 |
| 11 | Baffle tie/tie rod loose or damage | E1 CHT2, volt2, E1 CHT3 |
| 12 | Engine idle/RPM issue | volt2 |
| 14 | Engine need repair/reinstall/clean | volt1, amp1 |
| 15 | Engine failure/fire/time out | E1 CHT2 |
| 16 | Engine/propeller overspeed or damage | amp1 |
| 17 | Cylinder compression issue | E1 CHT2 |
| 18 | Pilot/in-flight noticed issue | E1 CHT2 |

The complementarity between different methods is equally evident — integrated gradients tends to identify contributions from exhaust gas temperatures (EGT series), while occlusion sensitivity focuses more on current and voltage signals — combining them yields a more comprehensive diagnostic explanation. From the fault type perspective, core engine faults (Categories 0, 1, 5, 14) commonly feature primary current and cylinder head temperature as consensus key sensors, while baffle-related faults (Categories 7, 10, 12) feature voltage signals as consensus key sensors, reflecting the differences in sensor responses arising from different fault mechanisms. The noise perturbation experiments also provide a complementary validation perspective for consistency: the key sensors unanimously identified by all methods under high confidence are precisely the channels showing the most significant attribution degradation after noise perturbation, i.e., the channels on which the model depends most heavily, corroborating the conclusions from the multi-sample statistical analysis.

In summary, the comprehensive interpretability analysis reveals that: primary battery current is the most universally applicable fault indicator; cylinder head temperatures and exhaust gas temperatures have significant differentiated diagnostic value for specific fault types; fault signatures are primarily concentrated in the mid-to-late flight segments, though mechanical structural faults such as cylinder cracks, compression issues, and overspeed can be detected in the early flight phase; noise perturbation experiments validate the reliability of attribution results from a causal relationship perspective. These findings provide maintenance personnel with clear directions for fault troubleshooting.

## 5 Discussion

This section provides an in-depth discussion of the experimental results from five perspectives: architecture design, lightweight strategy, knowledge distillation, cascaded system characteristics, and interpretability. It analyzes the advantages and limitations of the proposed method and identifies future research directions.

### 5.1 Engineering Value and System Characteristics of the Two-Stage Cascaded Architecture

The two-stage cascaded design decouples the optimization objectives of fault detection and fault identification, enabling the system to filter out the majority of normal samples with high recall in the first stage and concentrate the computational resources of the second stage on fine-grained discrimination among fault categories. This hierarchical workflow of "monitoring–alerting–diagnosis–decision-making"

naturally aligns with standard aviation maintenance practices and offers stronger engineering deployability compared to existing single-stage end-to-end methods. The modularity of each stage within the cascaded architecture also reduces system maintenance costs — when new fault types are introduced, only the second-stage model needs to be retrained without modifying the fault detection module.

From a system-level perspective, two aspects of the cascaded architecture's end-to-end characteristics merit attention. First, the overall missed detection rate is determined by the first stage. Faulty samples not detected by Stage 1 will completely escape and never enter Stage 2 for identification. Therefore, the recall of Stage 1 constitutes the safety bottleneck of the entire system. The distillation strategy in this work is currently applied to the Stage 2 fault identification task, and extending the distillation mechanism to Stage 1 to further improve the recall of anomaly detection is a worthwhile future direction. Under the current framework, Stage 1 can trade off between Precision and Recall by adjusting the binary classification threshold to accommodate deployment requirements at different safety levels. Second, computational resources are allocated on demand. Let $p$ denote the proportion of normal samples; the expected computational cost of the cascaded architecture is $C_1 + (1-p) \cdot C_2$, compared to $C_{full}$ for a single-stage method processing all samples, yielding more significant savings when the proportion of normal samples is high. In the dataset used in this work, $p \approx 0.5$, so the computational savings are relatively limited; however, in real-world operational environments, the vast majority of aircraft flights are in normal condition, $p$ is typically much greater than 0.5, and the efficiency advantage of the cascaded architecture becomes more pronounced. Nevertheless, this work currently reports only the independent performance of the two stages separately, and has not yet conducted an end-to-end evaluation of the two stages in series on a unified test set. The error propagation characteristics between the two stages — for example, the behavioral patterns of Stage 1 false positive samples upon entering Stage 2 — remain to be experimentally verified, representing an important direction for future work.

## 5.2 Design Insights from the LiteInception Lightweight Architecture

The 1+1 branch LiteInception achieves over 96.5% F1 retention with approximately 30% of the parameters, and the more than 8× CPU inference speedup is particularly critical for edge scenarios lacking GPU resources. The ablation experiments reveal two findings of reference value for lightweight architecture design in the time series domain.

First, the MaxPool branch is indispensable for capturing temporal extremal features. Removing MaxPool causes an abrupt F1 drop of 5 percentage points, far exceeding the impact of removing any single convolutional branch. This indicates that in aviation sensor signals, extremal features (such as temperature spikes and pressure drops) carry a higher density of diagnostic information than trend features, and the extraction capability of MaxPool for such information cannot be fully substituted by convolution operations.

Second, the marginal benefit of multi-scale convolutional branches is limited. The improvement from 3+1 to 2+1 is only 0.80 percentage points in F1, suggesting that under the current data characteristics, features extracted by convolutional branches with different kernel sizes exhibit high redundancy. Combined with the kernel size sensitivity experiment ($k = 3$ being optimal), for aviation sensor data resampled to 2,048 time steps, the strategy of small kernels with deep stacking achieves a better balance between receptive field coverage and nonlinear expressiveness, outperforming single-layer large-kernel mappings.

The above findings provide a valuable pathway for lightweight design in time series fault diagnosis: in aviation sensor signals, the complementary combination of single-scale convolution and pooling operations can effectively represent fault features, and multi-scale parallel designs are not essential. This conclusion may generalize to some extent to other industrial time series diagnosis scenarios with similar signal characteristics, but requires validation on additional datasets.

## 5.3 Knowledge Distillation as a Precision-Recall Adjustment Mechanism

The distillation experiments reveal a phenomenon that has not received sufficient attention previously: knowledge distillation serves not only as a means of model compression but also as a systematic mechanism for adjusting the Precision-Recall trade-off. The distilled version improves Recall by approximately 2.4 percentage points while decreasing Precision by approximately 1.5 percentage points, whereas the directly trained version exhibits the opposite characteristics. This phenomenon is consistent with the decision boundary smoothing effect of soft labels: the smoothed probability distribution from the teacher output at high temperature makes the student model more inclined to make classification decisions for ambiguous samples, thereby improving recall but also introducing more false positives.

This property enables the same network architecture to adapt to different deployment requirements simply by switching the training strategy, offering unique practical value in safety-critical domains. In general aviation maintenance scenarios, the consequence of missing a potential fault far outweighs that of issuing a false alarm, making the distilled version more suitable for automated alerting systems; conversely, in assisted diagnosis scenarios where experienced maintenance engineers perform the final interpretation, the high Precision of the directly trained version can reduce unnecessary troubleshooting workload. To the best of our knowledge, proposing and validating the use of knowledge distillation as a Precision-Recall adjustment tool rather than a pure compression technique is novel in the fault diagnosis domain.

It should be noted that the adjustment range of distillation on the Precision-Recall trade-off is constrained by the tunable ranges of the temperature coefficient and distillation weight, with the maximum Recall improvement in the current experiments being approximately 2.4 percentage points. In scenarios with extreme recall requirements (e.g., Recall > 95%), complementary approaches such as classification threshold adjustment or cost-sensitive learning may also be needed.

## 5.4 Physical Consistency of Interpretability Analysis

In the interpretability analysis, four attribution methods consistently identify the primary battery current sensor as the most universally applicable fault indicator, while cylinder head temperature and exhaust gas temperature sensors exhibit differentiated responses for specific fault types. These findings are highly consistent with the thermodynamic and electrical characteristics of the Cessna 172 piston engine — current directly reflects engine load variations, exhaust gas temperature is sensitive to combustion anomalies, and cylinder head temperature reflects the thermal equilibrium state. Multi-method cross-validation enhances the credibility of the interpretability results, addressing the common limitation in existing studies that rely on a single XAI method.

Notably, the interpretability analysis results also provide independent post-hoc validation for the channel selection decisions. The 15 sensor channels selected in Section 4.2 based on mutual information, gradient analysis, and SE attention weights highly overlap with the high-importance sensors identified by attribution methods in Section 4.6 (e.g., primary battery current, cylinder head temperature, and exhaust gas temperature series), demonstrating that the channel selection strategy indeed retains the features most contributive to model decisions. The conclusions from these two independent analysis steps corroborate each other.

## 5.5 Comparison with Existing Work

Placing the results of this work in the broader literature context further clarifies the positioning of the proposed method. The convolutional variational autoencoder method by Memarzadeh et al., based on the same NGAFID dataset, achieved flight fault detection in an unsupervised manner but did not address fine-grained fault identification and lacked interpretable outputs. The two-stage framework in this work provides fault detection while further achieving supervised classification of 19 fault categories, with dual-layer interpretability analysis providing evidential support for maintenance decisions. In terms of lightweight design, existing lightweight time series classification studies predominantly target single-device vibration signals and have not systematically explored simplification strategies for multi-branch architectures such as InceptionTime. The 1+1 branch simplification scheme proposed in this work and its ablation validation provide the first systematic benchmark for this direction.

The experimental results also reveal a clear architectural selection insight: for long-sequence, multi-channel time-domain signals such as those from general aviation sensors, the local feature extraction paradigm based on convolution substantially outperforms the global interaction paradigm based on attention. In the 19-class fault identification task, hybrid architectures containing Transformer components (CNN-Transformer, ConvMHSA) perform dramatically worse than pure convolutional architectures, and the pure Transformer almost completely fails. This stands in stark contrast to the dominance of Transformers in natural language processing and computer vision in recent years, suggesting that in industrial time series scenarios with limited data scale and strong local pattern characteristics, the inductive bias of convolutional architectures still holds significant advantages.

## 5.6 Limitations and Future Work

Despite achieving positive results, this study has the following limitations.

First, dataset singularity. The experiments are validated solely on the NGAFID dataset from a single aircraft type (Cessna 172), and the transferability to other propulsion types (turboprop, turbofan engines) and aircraft models remains to be investigated. The MSSP review guidelines emphasize the importance of multi-dataset validation, and future work will incorporate data from additional aircraft types to enhance the generalizability of the conclusions.

Second, absence of end-to-end cascaded evaluation. The current work reports the independent performance of the two stages separately but has not yet evaluated the end-to-end metrics of the two stages in series within a unified test pipeline (e.g., overall missed detection rate, cascaded error propagation characteristics). Future work needs to construct a complete cascaded evaluation pipeline to quantitatively analyze the impact of Stage 1 false positives and false negatives on Stage 2 outputs.

Third, lack of open-set recognition capability. The current system covers only 19 known fault categories and lacks the ability to detect unknown fault types, which in actual operations could lead to novel faults being misclassified as known categories.

Fourth, limitations of offline diagnosis. The system is positioned for post-flight offline diagnosis and does not yet support real-time streaming monitoring during flight.

Future research will proceed in the following directions: (1) exploring cross-aircraft transfer learning based on the pre-training and fine-tuning paradigm to reduce the data requirements for deploying on new aircraft types; (2) constructing an end-to-end evaluation framework for the two-stage cascade to systematically analyze error propagation characteristics and optimize the cascading strategy; (3) introducing open-set recognition mechanisms to enable the system to detect unknown fault types; (4) extending the framework to an online streaming architecture with sliding windows, achieving the capability upgrade from post-hoc diagnosis to in-flight early warning.

## 6 Conclusion

This paper proposes the LiteInception framework, providing a systematic solution across four dimensions — architecture design, model lightweight design, training strategy, and interpretability — for edge deployment requirements in general aviation fault diagnosis.

The two-stage cascaded architecture decouples fault detection from fault identification, aligns with standard maintenance workflows, and achieves on-demand allocation of computational resources. The 1+1 branch LiteInception compresses both the parameter count and computational cost of InceptionTime by 70%, accelerates CPU inference speed by more than 8×, while constraining performance loss to within 3%. The multi-method fusion-based channel selection strategy reduces the input dimensionality from 23 to 15 sensor channels, improving diagnostic performance while lowering computational overhead. The knowledge distillation strategy enables the same lightweight model to flexibly adjust Precision and Recall by switching training modes, adapting to deployment scenarios at different safety levels. The dual-layer interpretability framework, through cross-validation of four complementary attribution methods, provides a fault evidence chain traceable to specific sensors and temporal segments.

Experimental validation on the NGAFID dataset confirms the effectiveness of the above designs: the fault detection stage achieves 81.92% accuracy and 83.24% recall, while the fault identification stage achieves 77.00% accuracy. Future work will focus on extending cross-aircraft transfer learning and online streaming diagnosis capabilities.

## CRediT authorship contribution statement

Zhihuan Wei: Methodology, Software, Validation, Formal analysis, Investigation, Data curation, Visualization, Writing – original draft. Xinhang Chen: Software, Data curation, Formal analysis, Writing – original draft. Danyang Han: Investigation, Data curation, Validation. Yang Hu: Conceptualization, Methodology, Supervision, Funding acquisition, Writing – review & editing. Xuewen Miao: Conceptualization, Resources, Writing – review & editing. Guijiang Li: Conceptualization, Resources, Writing – review & editing.

## Declaration of competing interest

The authors declare that they have no known competing financial interests or personal relationships that could have appeared to influence the work reported in this paper.

## Declaration of Generative AI and AI-assisted technologies in the writing process

During the preparation of this work, the authors used DeepSeek in order to improve language and readability. After using this tool, the authors carefully reviewed and edited the content as needed and take full responsibility for the content of the publication.

## Acknowledgments

The work of Yang Hu was supported by the research funding of Hangzhou International Innovation Institute of Beihang University (Grant No. 2024KQ084)